\documentclass[graybox,envcountsect]{svmult}

\usepackage{mathptmx}       
\usepackage{helvet}         
\usepackage{courier}        
\usepackage{type1cm}        
\usepackage{makeidx}         
\usepackage{graphicx}        
\usepackage{multicol}        
\usepackage[bottom]{footmisc}

\usepackage{amssymb}
\usepackage{url}
\usepackage{verbatim}
\usepackage[numbers,sort&compress]{natbib}
\usepackage{dsfont}

\usepackage{amsmath}
\usepackage{amssymb}
\usepackage[svgnames]{xcolor}
\usepackage{hyperref}
\hypersetup{
  breaklinks,
  colorlinks,
  citecolor={DarkBlue},
  urlcolor={DarkBlue},
  linkcolor={green!50!black}
}
\usepackage[nameinlink]{cleveref}
\usepackage[textwidth=.9\marginparwidth,
  backgroundcolor=white,
  linecolor=gray,
  textsize=scriptsize,
  colorinlistoftodos]{todonotes}

\usepackage{etoolbox}
\makeatletter
\patchcmd{\NAT@test}{\else \NAT@nm}{\else \NAT@hyper@{\NAT@nm}}{}{}
\makeatother

\renewcommand{\E}{\mathbb{E}}
\newcommand{\R}{\mathds{R}}

\newcommand{\V}{\mathbb{V}}
\newcommand{\BigO}{\mathcal{O}}

\usepackage{mathtools}
\DeclarePairedDelimiter\ceil{\lceil}{\rceil}

\usepackage{pifont}

\usepackage{enumitem,amssymb}
\newlist{todolist}{itemize}{2}
\setlist[todolist]{label=\(\square\)}

\usepackage{algorithm}
\usepackage{algorithmic}

\newtheorem{assumption}{Assumption}
\crefname{assumption}{assumption}{assumptions}

\smartqed

\makeindex             

\usepackage[toc,page]{appendix}

\begin{document}

\title*{Adaptive First- and Second-Order Algorithms for Large-Scale Machine Learning}

\author{Sanae Lotfi, Tiphaine Bonniot de Ruisselet, Dominique Orban, Andrea Lodi}

\institute{Sanae Lotfi \at Department of Mathematical and Industrial Engineering, Polytechnique Montreal, Montreal, Quebec, Canada, \email{sanae.lotfi@polymtl.ca}
\and 
Tiphaine Bonniot de Ruisselet \at Department of Mathematical and Industrial Engineering, Polytechnique Montreal, Montreal, Quebec, Canada, \email{tiphaine.bonniot@gmail.com}
\and
Dominique Orban \at Department of Mathematics and Industrial Engineering, Polytechnique Montreal, Montreal, Quebec, Canada, \email{dominique.orban@polymtl.ca}
\and
Andrea Lodi \at Department of Mathematics and Industrial Engineering, Polytechnique Montreal, Montreal, Quebec, Canada,  \email{andrea.lodi@polymtl.ca}
}

\maketitle

\abstract{In this paper, we consider both first- and second-order techniques to address continuous optimization problems arising in machine learning. In the first-order case, we propose a framework of transition from deterministic or semi-deterministic to stochastic quadratic regularization methods. We leverage the two-phase nature of stochastic optimization to propose a novel first-order algorithm with adaptive sampling and adaptive step size. In the second-order case, we propose  a novel stochastic damped L-BFGS method that improves on previous algorithms in the highly nonconvex context of deep learning. Both algorithms are evaluated on well-known deep learning datasets and exhibit promising performance.
}


\medskip


\section{Introduction and related work}%
\label{sec:intro}

In this paper, we explore promising research directions to improve upon existing first- and second-order optimization algorithms in both  deterministic and stochastic settings, with special emphasis on solving problems that arise in the machine learning context. 

We consider general unconstrained stochastic optimization problems of the form
\begin{equation}
  \label{eq:optim-prob1}
  \min_{x \in \R^n} \E_\xi[F(x,\xi)],
\end{equation}
where \(\xi \in \R^d\) denotes a random variable and \(F : \R^n \times \R^d \to \R\) is continuously differentiable and may be nonconvex. We introduce additional assumptions about \(F\) as needed. We refer to~\eqref{eq:optim-prob1} as an \emph{online} optimization problem, which indicates that we typically sample data points during the optimization process so as to use \emph{new} samples at each iteration, instead of using a fixed dataset that is known up-front. We refer to \(\E_\xi[F(x,\xi)]\)
as the \emph{expected risk} or the \emph{expected loss}.

In machine learning, a special case of~\eqref{eq:optim-prob1} is \emph{empirical risk minimization}, which consists in solving
\begin{equation}
  \label{eq:stoch-optim-prob}
  \min_{x \in \R^n} f(x), \qquad
  f(x) := \frac{1}{N} \sum_{i=1}^{N} f_i(x) , 
\end{equation}
where \(f_i\) is the loss function that corresponds to the \(i\)-th element of our dataset and \(N\) is the number of samples, i.e., the size of the dataset. We refer to~\eqref{eq:stoch-optim-prob} as a \emph{finite-sum} problem and to \(f\) as the \emph{empirical risk} or the \emph{empirical loss}.

Stochastic Gradient Descent (SGD) \citep{robbins1951stochastic, bottou2010large} and its variants \citep{polyak1964some, nesterov1983method, duchi2011adaptive, tieleman2012rmsprop, Adam}, including variance-reduced algorithms \citep{svrg,nguyen2017sarah,fang2018spider, wang2019spiderboost}, are among the most popular methods for~\eqref{eq:optim-prob1} and~\eqref{eq:stoch-optim-prob} in machine learning. They are the typical benchmark with respect to developments in both first- and second-order methods.

In the context of first-order methods, we focus on the case where it is not realistic to evaluate \(\nabla f(x)\) exactly; rather, we have access to an approximation \( g(x_k, \xi_k)\). If we consider~\eqref{eq:stoch-optim-prob}, a popular choice for \( g(x_k, \xi_k)\) is the \emph{mini-batch} gradient
\begin{equation}
  \label{eq:minibatch-gradient}
  g(x_k, \xi_k) = \frac{1}{m_k} \sum_{i=1}^{m_k} \nabla f_{\xi_{k,i}}(x_k),
\end{equation}
where \(\xi_{k}\) is the subset of samples considered at iteration \(k\), from a given set of realizations of \(\xi\), \(\xi_{k,i}\) is the \(i\)-th sample of \(\xi_{k}\) and \(m_k\) is the batch size used at iteration \(k\), i.e., the number of samples used to evaluate the gradient approximation. It follows that \(f_{\xi_{k,i}}\) represents the loss function that corresponds to the sample \(\xi_{k,i}\).

As discussed previously, 
stochastic optimization methods are the default choice for solving large-scale optimization problems. 
However, the search for the most problem-adapted hyperparameters involves significant computational costs. \citet{asi2019importance} highlighted the alarming computational and engineering energy used to find the best set of hyperparameters in training neural networks by citing 3 recent works \citep{real2019regularized, zoph2016neural, collins2016capacity}, where the amount of time needed to tune the optimization algorithm and find the best neural structure can be equivalent to up to \(750,000\) central processing unit days \citep{collins2016capacity}. 
Therefore, we need to design optimization algorithms with adaptive parameters, that automatically adjust to the nature of the problem without the need for a hyperparameter search. 
\citet{larson2016stochastic, chen2018stochastic, blanchet2019convergence}, and \citet{curtis2019stochastic} proposed stochastic methods with adaptive step size using the trust-region framework. The use of an adaptive batch size throughout the optimization process is another promising direction towards adaptive algorithms. Exiting works include \citet{friedlander2012hybrid, byrd2012sample, hashemi2014adaptive}, and \citet{bollapragada2018adaptive}. We propose a novel algorithm with both adaptive step size and batch size in \Cref{sec:firstorder}. 

Another potential area of improvement is in the context of solving machine learning problems that are highly nonconvex and ill-conditioned \citep{bottou2018optimization}, which are more effectively treated using (approximate) second-order information.
Second-order algorithms are well studied in the deterministic case \citep{dennis1974characterization, dembo1982inexact, dennis1996numerical, amari1998natural} but there are many areas to explore in the stochastic context that go beyond existing works \citep{schraudolph2007stochastic, bordes2009sgd, byrd2016stochastic, moritz2016linearly,gower2016stochastic}.
Among these areas, the use of damping in L-BFGS is an interesting research direction to be leveraged in the stochastic case. Specifically, \citet{wang2017stochastic} proposed a stochastic damped L-BFGS (SdLBFGS) algorithm and proved almost sure convergence to a stationary point.
However, damping does not prevent the inverse Hessian approximation \(H_k\) from being ill-conditioned \citep{chen2019stochastic}.
The convergence of SdLBFGS may be heavily affected if the Hessian approximation becomes nearly singular during the iterations.
In order to remedy this issue, \citet{chen2019stochastic} proposed to combine SdLBFGS with regularized BFGS \citep{mokhtari2014res}.
The approach we propose in \Cref{sec:secondorder} differs.

\subsubsection*{Notation}

We use \(\|.\| := \|.\|_2\) to denote the Euclidean norm. Other types of norms will be introduced by the notation \(\|.\|_p\), where the value of \(p\) will be specified as needed. Capital Latin letters such as \(A\), \(B\), and \(H\) represent matrices, lowercase Latin letters such as \(s\), \(x\), and \(y\) represent vectors in \(\R^n\), and lowercase Greek letters such as \(\alpha\), \(\beta\) and \(\gamma\) represent scalars. The operators \(\E_\xi[.]\) and \(\V_\xi[.]\) represent the expectation and variance over random variable \(\xi\), respectively. For a symmetric matrix \(A\), we use \(A \succ 0\) and \(A \succeq 0\) to indicate that \(A\) is, respectively, positive definite or positive semidefinite, and \(\lambda_{\min}(A)\) and \(\lambda_{\max}(A)\) to denote its smallest and largest eigenvalue, respectively.
The identity matrix of appropriate size is denoted by \(I\). We use \(\ceil*{.}\) to denote the ceiling function that maps a real number \(y\) to the smallest integer greater than or equal to \(y\).

\subsection{Framework and definitions} 

We consider methods in which iterates \(x_k\) are updated as follows: 
\begin{equation}
  \label{eq:theta}
  x_{k+1} = x_k + \alpha_k d_k,
  \quad
  d_k = -B_k^{-1}  g(x_k, \xi_k),
  \quad
  (k \geq 0),
\end{equation}
where \(\alpha_k > 0\) is the step size, \(d_k\) is the search direction,
\(B_k = B_k^T \approx \nabla^2 f(x_k)\) is positive definite, and we set \(H_k = B_k^{-1}\).
Setting \(B_k\) to a multiple of \(I\) results in a first-order method.

A common practice in machine learning is to divide the dataset into a \emph{training} set, that is used to train the chosen model, and a \emph{test} set, that is used to evaluate the performance of the trained model on unseen data. The training and test sets are usually partitioned into \emph{batches}, which are subsets of these sets. The term \emph{epoch} refers to an optimization cycle over the entire training set, which means that all batches are used during this cycle. 

We use the expression \emph{semi-deterministic approach} to refer to an optimization algorithm that uses exact function values and inexact gradient values. 

\subsection{Contributions and organization}

The paper is naturally divided in two parts.

\Cref{sec:firstorder} presents our contributions to first-order optimization methods wherein we define a framework to transform a semi-deterministic quadratic regularization method, i.e., one that uses exact function values and inexact gradient estimates, to a fully stochastic optimization algorithm suitable for machine learning problems. 
We go a step further to propose a novel first-order optimization algorithm with adaptive step size and sampling that exploits the two-phase nature of stochastic optimization. 
The result is a generic framework to adapt quadratic regularization methods to the stochastic setting. 
We demonstrate that our implementation is competitive on machine learning problems.

\Cref{sec:secondorder} addresses second-order methods. We propose a novel stochastic damped L-BFGS method.
An eigenvalue control mechanism ensures that our Hessian approximations are uniformly bounded using less restrictive assumptions than related works.
We establish almost sure convergence to a stationary point, and demonstrate that our implementation is more robust than a previously proposed stochastic damped L-BFGS algorithm in the highly nonconvex case of deep learning.

Finally, \Cref{sec:final} draws conclusions and outlines potential research directions.

\section{Stochastic adaptive regularization with dynamic sampling}%
\label{sec:firstorder}

This section is devoted to first-order methods and provides a framework of transition from deterministic or semi-deterministic to fully stochastic quadratic regularization methods. In \Cref{sec:arig}, we consider the generic unconstrained problem
\begin{equation}
  \label{eq:unc}
  \min_{x \in \R^n} \ f(x),
\end{equation}
where \(f: \R^n \to \R\) has Lipschitz-continuous gradient.
We assume that the gradient estimate accuracy can be selected by the user, but do not require errors to converge to zero. 
We propose our Adaptive Regularization with Inexact Gradients (ARIG) algorithm and establish global convergence and optimal iteration complexity. 

In \Cref{sec:stochastic-arig}, we specialize our findings to a context appropriate for machine learning, and consider online optimization problems in the form~\eqref{eq:optim-prob1} and~\eqref{eq:stoch-optim-prob}.
We prove that we can use inexact function values in ARIG without losing its convergence guarantees.

We present a stochastic variant of ARIG, as well as a new stochastic adaptive regularization algorithm with dynamic sampling that leverages \citeauthor{pflug1990non}'s diagnostic in \Cref{sec:new-1st-order-stoch}. We evaluate our implementation on machine learning problems in \Cref{sec:p1-test}.

\subsection{Adaptive regularization with inexact gradients (ARIG)}\label{sec:arig}

We consider~\eqref{eq:unc} and assume that it is possible to obtain \(\nabla f(x, \omega_g) \approx \nabla f(x)\) using a user-specified relative error threshold \(\omega_g > 0\), i.e.,

\begin{equation}
  \label{eq:g-error}
  \|\nabla f(x, \omega_g) - \nabla f(x, 0)\| \leq \omega_g \|\nabla f(x, \omega_g)\|,
  \quad \text{with} \ \nabla f(x, 0) = \nabla f(x).
\end{equation}
Geometrically,~\eqref{eq:g-error} requires the exact gradient to lie within a ball centered at the gradient estimate.

%
%
%
%
%
%
%



We define our assumptions as follows:



\begin{assumption}%
  \label{asm:f-C1}
  The function \(f\) is continuously differentiable over \(\R^n\).
\end{assumption}

\begin{assumption}%
  \label{asm:g-lipschitz}
  The gradient of \(f\) is Lipschitz continuous, i.e., there exists \(L > 0\) such that for all \(x\), \(y \in \R^n\), \(\|\nabla f(x) - \nabla f(y)\| \leq L \, \|x - y\|\).
\end{assumption}

\subsubsection{Formulation for ARIG}

We define $T(x,s)$ as follows:
\begin{equation*}
  T(x,s) := f(x) + \nabla {f(x)}^\top s \approx f(x+s).
\end{equation*}
%
%
%
%
%
%
At iteration \(k\), we consider the approximate Taylor model using the inexact gradient \(\nabla f(x, \omega_g)\) defined in~\eqref{eq:g-error}:
\begin{equation}
\label{eq:tayerror}
  \bar{T}_k (s) := f(x_k) + \nabla {f(x_k,\omega_g^k)}^\top  s \approx T(x_k, s).
\end{equation}

Finally, we define the approximate regularized Taylor model
\begin{equation}
  \label{eq:model}
  l_k(s) := \bar{T}_k(s) + \tfrac{1}{2} \sigma_k \|s\|^2,
\end{equation}
whose gradient is
\begin{equation*}
  \nabla_s l_k(s)= \nabla_s \bar{T}_k(s) +\sigma_k s = \nabla f(x_k,\omega_g^k) + \sigma_k s,
\end{equation*}
where \(\sigma_k > 0\) is the iteration-dependent regularization parameter.



\Cref{alg:regularization-inexact} summarizes our Adaptive Regularization with Inexact Gradients method.
The approach is strongly inspired by the framework laid out by \citet{birgin-gardenghi-martinez-santos-toint-2017} and the use of inexact gradients in the context of trust-region methods proposed by \citet{carter-1991} and described by \citet[\S8.4]{conn-gould-toint-2000}.
The adaptive regularization property is apparent in step~\ref{step5:alg:regularization-inexact} of \Cref{alg:regularization-inexact}, where \(\sigma_k\) is updated based on the quality of the model \(l_k\).
\(\sigma_k\) can be interpreted as the inverse of a trust-region radius or as an approximation of \(L\).
The main insight is that the trial step \(s_k\) is accepted if the ratio
\begin{equation*}
  \rho_k =
  \frac{f(x_k)-f(x_k+s_k)}{\bar{T}_k(0)-\bar{T}_k(s_k)} =
  \frac{f(x_k)-f(x_k+s_k)}{\frac{1}{\sigma_k}\|\nabla f(x_k,\omega_g^k)\|^2},
\end{equation*}
of the actual objective decrease to that predicted by the model is sufficiently large.
The regularization parameter is then updated using the following scheme:
\begin{itemize}
  \item If \(\rho_k \geq \eta_2\), then the model quality is sufficiently high that we decrease \(\sigma_k\) so as to promote a larger step at the next iteration;
  \item if \( \eta_1 \leq \rho_k < \eta_2\), then the model quality is sufficient to accept the trial step \(s_k\), but insufficient to decrease \(\sigma_k\) (in our implementation, we increase \(\sigma_k\));
  \item if \( \rho_k < \eta_1\), then the model quality is low and \(s_k\) is rejected. We increase \(\sigma_k\) more aggressively than in the previous case so as to compute a conservative step at the next iteration.
\end{itemize}



\begin{algorithm}[htbp]
  \caption{Adaptive Regularization with Inexact Gradients---ARIG}%
  \label{alg:regularization-inexact}
  \begin{algorithmic}[1]

    \STATE%
    \label{step1:alg:regularization-inexact}
    Choose a tolerance \(\epsilon > 0\), the initial regularization parameter \(\sigma_0 > 0\), and the constants 
    \begin{equation}
      \label{ar:parameters1}
      0 < \sigma_{\min} \leq \sigma_0, \quad
      0 < \eta_1 \leq \eta_2 < 1 \quad \text{and} \quad
      0 < \gamma_1 < 1 < \gamma_2 < \gamma_3.
    \end{equation}
    Set \(k = 0\).

    \STATE%
    \label{step2:alg:regularization-inexact}
    Choose \(\omega_g^k\) such that \(0 < \omega_g^k \leq 1 / \sigma_k\) and compute \(\nabla f(x_k,\omega_g^k)\) such that~\eqref{eq:g-error} holds.
    If \(\|\nabla f(x_k,\omega_g^k)\|\leq \epsilon / (1+\omega_g^k)\), terminate with the approximate solution \(x_{\epsilon} = x_k\).

    \STATE%
    \label{step3:alg:regularization-inexact}
    Compute the step \(s_k = -\frac{1}{\sigma_k}\nabla f(x_k,\omega_g^k)\).
    
    \STATE%
    \label{step4:alg:regularization-inexact}
    Evaluate \(f(x_k+s_k)\) and define
    \begin{equation}
      \label{ar:ratio}
      \rho_k = \frac{f(x_k)-f(x_k+s_k)}{\bar{T}_k(0)-\bar{T}_k(s_k)} = \frac{f(x_k)-f(x_k+s_k)}{\frac{1}{\sigma_k}\|\nabla f(x_k,\omega_g^k)\|^2}.
    \end{equation}
    If \(\rho_k \geq \eta_1\), set \(x_{k+1} = x_k + s_k\).
    Otherwise, set \(x_{k+1} = x_k\).

    \STATE%
    \label{step5:alg:regularization-inexact}
    Set
    \begin{equation}
      \label{ar:update}
      \sigma_{k+1} \in
      \begin{cases}
        [\max(\sigma_{\min}, \gamma_1\sigma_k), \sigma_k]
        &\text{ if } \rho_k \geq \eta_2, \\
        [\sigma_k, \gamma_2\sigma_k]
        &\text{ if } \eta_1 \leq \rho_k < \eta_2, \\
        [\gamma_2\sigma_k, \gamma_3\sigma_k]
        &\text{ if } \rho_k < \eta_1. \\
      \end{cases}
    \end{equation}
    Increment \(k\) by one and go to step~\ref{step2:alg:regularization-inexact} if \(\rho_k \geq \eta_1\) or to step~\ref{step3:alg:regularization-inexact} otherwise.
  \end{algorithmic}
\end{algorithm}


\subsubsection{Convergence and complexity analysis}\label{sec:convergence}

The condition~\eqref{eq:g-error} ensures that at each iteration \(k\),
\begin{equation*}
  \|\nabla f(x_k)\| \leq
  \|\nabla f(x_k)-\nabla f(x_k,\omega_g^k)\| + \|\nabla f(x_k,\omega_g^k)\| \leq
  (1+\omega_g^k) \|\nabla f(x_k,\omega_g^k)\|.
\end{equation*}
Therefore, when the termination occurs, the condition \(\|\nabla f(x_k,\omega_g^k)\| \leq \epsilon / (1+\omega_g^k)\)  implies \(\|\nabla f(x_\epsilon)\| \leq \epsilon\) and \(x_\epsilon\) is an approximate first-order critical point to within the desired stopping condition.

The following analysis represents the adaptation of the general properties presented by \citep{birgin-gardenghi-martinez-santos-toint-2017} to a first-order model with inexact gradients.

We first obtain an upper bound on \(\sigma_k\).
The following result is similar to \citet[Lemma~2.2]{birgin-gardenghi-martinez-santos-toint-2017}.
All proofs may be found in \Cref{sec-appendix}.

\begin{lemma}%
  \label{lem:sigma-bounded}
  For all \(k \geq 0\),
  \begin{equation}
    \label{eq:sigma-max}
    \sigma_k \leq \sigma_{\max} := \max
    \left[\sigma_0, \frac{\gamma_3 (\tfrac{1}{2}L+1)}{1-\eta_2} \right].
  \end{equation}
\end{lemma}

We now bound the total number of iterations in terms of the number of successful iterations and state our main evaluation complexity result.




\begin{theorem}[{\protect\citealp[Theorem~\(2.5\)]{birgin-gardenghi-martinez-santos-toint-2017}}]%
  \label{thm:arig-complexity}
  Let \Cref{asm:f-C1,asm:g-lipschitz} be satisfied.
  Assume that there exists \(\kappa_{\text{low}}\) such that \(f(x_k) \geq f_{\text{low}}\) for all \(k \geq 0\).
  Assume also that \(\omega_g^k \leq 1/\sigma_k\) for all \(k \geq 0\).
  Then, given \(\epsilon > 0\), \Cref{alg:regularization-inexact} needs at most
  \begin{equation*}
    \left\lfloor\kappa_s \frac{f(x_0)-f_\mathrm{low}}{\epsilon^2}\right\rfloor
  \end{equation*}
  successful iterations (each involving one evaluation of \(f\) and its approximate gradient) and at most
  \begin{equation*}
    \left\lfloor
    \kappa_s \frac{f(x_0)-f_\mathrm{low}}{\epsilon^2}
    \right\rfloor
    \left( 1+\frac{|\log \gamma_1|}{\log \gamma_2} \right) +
    \frac{1}{\log\gamma_2}
    \log \left( \frac{\sigma_{\max}}{\sigma_0}\right)
  \end{equation*}
  iterations in total to produce an iterate \(x_{\epsilon}\) such that
  \(\|\nabla f(x_{\epsilon})\| \leq \epsilon, \ \text{where} \ \sigma_{\max}\) is given by \Cref{lem:sigma-bounded} and where
  \begin{equation*}
    \kappa_s = \frac{{(1+\sigma_{\max})}^2}{\eta_1\sigma_{\min}}.
  \end{equation*}
\end{theorem}

\Cref{thm:arig-complexity} implies that \(\liminf \|\nabla f(x_k)\| = 0\) provided \(\{f(x_k)\}\) is bounded below.
The following corollary results immediately.

\begin{corollary}
  Under the assumptions of \Cref{thm:arig-complexity}, either \(f(x_k) \to -\infty\) or \(\liminf \|\nabla f(x_k)\| = 0\).
\end{corollary}



\subsection{Stochastic adaptive regularization with dynamic sampling}\label{sec:stochastic-arig}

Although ARIG enjoys strong theoretical guarantees under weak assumptions, it is not adapted to the context of machine learning, where we do not have access to exact objective values. Therefore, we propose to build upon ARIG in order to design a stochastic first-order optimization with adaptive regularization with similar convergence guarantees.

\subsubsection{Adaptive regularization with inexact gradient and function values}\label{sec:newarig1}

\citet[\S10.6]{conn-gould-toint-2000} provide guidelines to allow for inexact objective values in trust-region methods while preserving convergence properties.
This section adapts their strategy to the context of \Cref{alg:regularization-inexact}.
Assume that we do not have access to direct evaluations of \(f(x)\), but we can obtain an approximation \(f(x_k, \omega_f^k)\) depending on a user-specified error threshold \(\omega_f^k\), such that
\begin{equation}
  \label{eq:f-error}
  |f(x, \omega_f^k) - f(x, 0)| \leq \omega_f^k,
  \quad \text{with} \quad \ f(x, 0) = f(x).
\end{equation}

We redefine the approximate Taylor series~\eqref{eq:tayerror} using inexact gradient and function values:
\begin{equation}%
  \label{eq:tayerror2}
  \bar{T}_k (s) = f(x, \omega_f^k) + \nabla {f(x_k,\omega_g^k)}^\top  s.
\end{equation}

If the approximations of \(f(x_k)\) and \(f(x_k + s_k)\) are not sufficiently accurate, the ratio \(\rho_k\) defined in~\eqref{ar:ratio} loses its meaning, since it is supposed to quantify the ratio of the \emph{\textbf{actual} decrease} of the function value to the \emph{predicted decrease} by the model.
To circumvent the difficulty, we have the following result whose proof is elementary.

\begin{lemma}%
  \label{lem:inexact-decrease}
  Let the constant \(\eta_0\) be such that \(\eta_0 < \frac{1}{2} \eta_1\) and let \(s_k\) be a step taken from \(x_k\) such that \(\bar{T}_k(s_k) < \bar{T}_k(0)\).
  If \(\omega_f^k\) and \(\hat{\omega}_f^k\) satisfy,
  \begin{equation}
    \label{eq:original-important-condition}
    \max(\omega_f^k, \hat{\omega}_f^k) \leq \eta_0  \left[\bar{T}_k(0)-\bar{T}_k(s_k)\right],
  \end{equation}
  then, a sufficient decrease in the inexact function values,
  \begin{equation*}
    \bar{\rho}_k := \frac{\bar{f}(x_k, \omega_f^k) - \bar{f}(x_k+s_k, \hat{\omega}_f^k)}{\bar{T}_k(0)-\bar{T}_k(s_k)} \geq \eta_1,
  \end{equation*}
  implies a sufficient decrease in the actual function values,
  \begin{equation*}
    \rho_k = \frac{f(x_k)-f(x_k+s_k)}{\bar{T}_k(0)-\bar{T}_k(s_k)} \geq \bar{\eta}_1, 
  \end{equation*}
  where \(\bar{\eta}_1 = \eta_1 - 2 \eta_0 > 0\). 
\end{lemma}

\Cref{lem:inexact-decrease} states that whenever~\eqref{eq:original-important-condition} is satisfied, a sufficient decrease in the \emph{stochastic} objective, i.e., \( \bar{\rho}_k \geq \eta_1\), implies that the \emph{actual} decrease in the exact objective is at least a fraction \(\bar{\eta}_1\) of the decrease predicted by the model. It also means that, as long as we can guarantee that~\eqref{eq:original-important-condition} is satisfied at each iteration, the convergence and complexity results from the previous section continue to apply.



In the next section, we use adaptive sampling to satisfy~\eqref{eq:g-error} in expectation with respect to the batch used to obtain \(\nabla f(x, \omega_g)\) without evaluation of the full gradient.

\subsubsection{Stochastic ARIG with adaptive sampling}\label{sec:newarig2}

To use a stochastic version of ARIG in machine learning, we must compute a gradient approximation \(g(x_k, \xi_k)\) for mini-batch \(\xi_k\),
\begin{equation}
\label{eq:av-grad}
 g(x_k, \xi_k) = \frac{1}{m_k} \sum_{i=1}^{m_k} g(x_k, \xi_{k,i}),
\end{equation}
that satisfies
\begin{equation}
  \label{eq:g-error2}
  \|g(x_k, \xi_k) - \nabla f(x_k)\| \leq \omega_g^k \|g(x_k, \xi_k)\|,
\end{equation}  
where \(\omega_g^k\) is an iteration-dependent error threshold. 


Condition~\eqref{eq:g-error2} is sometimes called the \emph{norm test}.
We can use the adaptive sampling strategy of \citet{byrd2012sample} to maintain this condition satisfied along the iterations.
Let us introduce their adaptive sampling strategy.
Since \(\omega_g^k\) should verify \(0 < \omega_g^k \leq 1 / \sigma_k\), let us fix it to \(\omega_g^k := 1 / \sigma_k\). This choice is natural since a higher error-threshold would require the use of fewer samples to compute \(g(x_k, \xi_k)\).
Since \(g(x_k, \xi_k)\) is obtained as a sample average~\eqref{eq:av-grad}, it is an unbiased estimate of \(\nabla f(x_k)\). Therefore,
\begin{equation}
  \label{eq:exp_grad_error_new}
  \mathbb{E}_{\xi_k}[\|g(x_k, \xi_k) - \nabla f(x_k) \|^2] = \| \V_{\xi_k}[g(x_k, \xi_k)] \|_1,
\end{equation}
where \(\V_{\xi_k}[g(x_k, \xi_k)]\) is a vector with components \(\V_{\xi_k}[g_j(x_k, \xi_k)]\), \(j = 1, \dots, n\).
\citet[p.183]{freund1971mathematical} establish that,
\begin{equation}
  \label{eq:var_approx12}
  \V_{\xi_k}[g(x_k, \xi_k)] = \frac{\V_{i \in \xi_k}(\nabla f_i(x_k))}{m_k} \, \frac{(N - m_k)}{N-1}.
\end{equation}

The direct computation of the population variance \(\V_{i \in \xi_k}(\nabla f_i(x_k))\) is expensive. Hence, \citet{byrd2012sample} suggest to estimate it with the sample variance
\begin{equation}
  \label{eq:var_approx2_new}
  \hat{\V}_{i \in \xi_k}\left(\nabla f_i(x_k)\right) = \frac{1}{m_k - 1} \sum_{i \in \xi_k} {(\nabla f_i(x_k) -  g(x_k, \xi_k))}^2,
\end{equation}
where the square is applied elementwise.
It follows from~\eqref{eq:exp_grad_error_new} and~\eqref{eq:var_approx12} that
\begin{equation}
  \label{eq:exp_grad_error2_new}
  \mathbb{E}_{\xi_k}\left[\|g(x_k, \xi_k) - \nabla f(x_k) \|^2\right] =
  \frac{\|\hat{\V}_{i \in \xi_k}(\nabla f_i(x_k))\|_1}{m_k} \, \frac{(N - m_k)}{N-1}.
\end{equation}

In the context of large-scale optimization, we let \(N \to \infty\), and~\eqref{eq:g-error2} implies
\begin{equation}
  \label{eq:error-grad2_new}
  \frac{\|\hat{\V}_{i \in \xi_k}\left(\nabla f_i(x_k)\right)\|_1}{m_k} \leq \frac{1}{\sigma_k^2} \|g(x_k, \xi_k)\|^2.
\end{equation}

Finally, our strategy to obtain \(g(x_k, \xi_k)\) that satisfies~\eqref{eq:g-error2}, with \(\omega_g^k = 1 / \sigma_k\), is as follows:
\begin{enumerate}
    \item at iteration \(k\), sample a new batch of size \(m_k = m_{k-1}\) and compute the sample variance \(\hat{\V}_{i \in \xi_k}\) and mini-batch gradient \(g(x_k, \xi_k)\);
    \item if~\eqref{eq:error-grad2_new} holds, use mini-batch \(m_k\) at iteration \(k\). Otherwise, sample a new batch \(\tilde{\xi}\) of size
    \begin{equation}
      \label{eq:new_batchsize1_new}
      \Tilde{m}_k := \ceil*{\frac{\sigma_k^2 \|\hat{\V}_{i \in \xi_k}(\nabla f_i(x_k))\|_1}{\|g(x_k, \xi_k)\|^2}}.
    \end{equation}
\end{enumerate}

It is important to note that the strategy above relies on the assumption that the change in the batch size is gradual.  Therefore, we assume that
\begin{equation}
  \label{assum:approx_new}
  \|\hat{\V}_{i \in \tilde{\xi}_k}\left(\nabla f_i(x_k)\right)\|_1 \approx \|\hat{\V}_{i \in \xi_k}(\nabla f_i(x_k))\|_1  \quad \text{and} \quad \|g(x_k, \Tilde{\xi}_k)\| \approx \|g(x_k, \xi_k)\|.
\end{equation}

\citet{lotfi2020stochastic} provides a comparison between stochastic ARIG with adaptive sampling and SGD on the MNIST dataset \citep{lecun2010mnist}. 







Theoretically speaking, convergence requires that either the variance of the gradient estimates or the step size decays to zero. For the strategy defined above, the step size does not converge to zero. 
Moreover, the percentage batch size would need to be close to \(100\%\) for the gradient variance to converge to zero. 
This discussion motivates our work in \Cref{sec:new-1st-order-stoch}, where we propose a new algorithm that leverages both adaptive regularization and adaptive sampling, but not simultaneously.
Although we do not provide theoretical guarantees, it is conceivable that the new algorithm globally converges to a stationary points since we choose a step size that converges to zero in the second phase of the optimization~\citep{bottou2018optimization}.

\subsection{Stochastic adaptive regularization with dynamic sampling and convergence diagnostics}\label{sec:new-1st-order-stoch}

In this section, we present a novel algorithm that incorporates both adaptive regularization and adaptive sampling by leveraging the two-phase nature of stochastic optimization.

\subsubsection{Statistical testing for stochastic optimization}

Stochastic optimization algorithms exhibit two phases: a transient phase and a stationary phase \citep{murata1998statistical}. The transient phase is characterized by the convergence of the algorithm to a region of interest that contains potentially a good local or global optimum. In the stationary phase, the algorithm oscillates around that optimum in the region of interest.

Several authors aimed to derive a stationary condition to indicate when stochastic procedures leave the transient phase and enter the stationary phase. Some of the early works in this line of research are Pflug's works \citep{pflug1983determination, pflug1990non, pflug1992gradient}. In particular, Pflug's procedure was introduced in \citet{pflug1990non}, and it consists of keeping a running average of the inner product of successive gradients in order to detect the end of the transient phase. Pflug's idea is simple: when the running average becomes negative, that suggests that the consecutive stochastic gradients are pointing in different directions. This should signal that the algorithm entered the stationary phase where it oscillates around a local optimum. More recently, \citet{chee2018convergence} developed a statistical diagnostic test in order to detect when Stochastic Gradient Descent (SGD) enters the stationary phase, by combining the statistical test from \citet{pflug1990non} with the SGD update. They presented theoretical arguments and practical experiments to prove that the test activation occurs in the phase-transition for SGD\@. \Cref{alg:pflug-diagnostic} describes Pflug's diagnostic proposed by \citet{chee2018convergence} for convergence of SGD applied to~\eqref{eq:stoch-optim-prob}, where \(g(x_k) \) represents an approximation to the full gradient \(\nabla f(x_k)\). Note that in this case, \(\xi_k\) contains a single sample.
\citeauthor{chee2018convergence} prove that the convergence diagnostic in \Cref{alg:pflug-diagnostic} satisfies \(\E(S_k - S_{k-1}) < 0\) as \(k \to +\infty\), which implies that the algorithm terminates almost surely. 

\begin{algorithm}[htbp]
  \caption{Pflug diagnostic for convergence of SGD}%
  \label{alg:pflug-diagnostic}
  \begin{algorithmic}[1]
  \REQUIRE%
  \(x_0 \in \mathbb{R}^n\)
  \STATE%
    Choose \(\gamma > 0\) and a positive integer named ``burn-in''.
  \STATE%
    Set \(S_0 = 0\).
    Compute \(g(x_0)\) and \(x_{1} = x_0 - \gamma g(x_0)\).
  \FOR{\(k = 0, 1, \ldots\)}
    \STATE%
    Compute \(g(x_k)\) and \(x_{k+1} = x_k - \gamma g(x_k)\).
    \STATE%
    Define \(S_k = S_{k-1} + {g(x_k)}^\top g(x_{k-1})\).
    \IF{\(k > \text{burn-in}\) and \(S_k < 0\)} 
      \STATE%
    return \(k\).
    \ENDIF
  \ENDFOR
  \end{algorithmic}
\end{algorithm}

\subsubsection{Adaptive regularization and sampling using Pflug diagnostic}

To harness Pflug's diagnostic, we discuss the bias-variance trade-off in machine learning. The bias of a model is the extent to which it approximates a global optimum,
whereas its variance is the amount of variation in the solution if the training set changes. The generalization error in machine learning is the sum of the bias and the variance, hence the trade-off between the two quantities. Now notice that
\begin{itemize}
    \item in the transient phase, we would like to reduce the bias of the model by converging to a promising region of interest. Therefore, we propose to use adaptive regularization in order to adapt the step size automatically, taking into account the ratio \(\rho_k\). We use a fixed batch size during this phase.
    \item In the stationary phase, we would like to reduce the variance of the gradient estimates. Therefore, we use adaptive sampling as a variance reduction technique. Additionally, we choose a step size sequence that converges to zero to ensure global convergence to a stationary point.
\end{itemize}

We call this algorithm ARAS for Adaptive Regularization and Adaptive Sampling algorithm. \Cref{alg:regularization-inexact_new2} states the complete procedure. Note that we reduce the number of parameters that were initially introduced in ARIG by eliminating \(\gamma_3\) and \(\eta_2\), which simplifies the algorithm. Inspired by \citet{curtis2019stochastic}, we do not require the decrease to be sufficient to accept a step. Instead, we accept the step in all cases. We also use a constant maximum batch size, \(m_{\max}\) to ensure that the batch sizes stay reasonable. 

\begin{algorithm}[htbp]
  \caption{Adaptive Regularization and Sampling using Pflug diagnostic---ARAS}%
  \label{alg:regularization-inexact_new2}
  \begin{algorithmic}[1]
    \STATE%
    \label{step3:alg:regularization-inexact_new2}
    Choose the initial regularization parameter \(\sigma_0 > 0\), the initial batch size \(m_0\), the maximum batch size \(m_{\max}\), the total number of epochs \(N_{\text{epochs}}\), \( \text{burn-in} > 0\) and the constants 
    \(0 < \sigma_{\min} \leq \sigma_0\),  \(0 < \eta < 1\), and \(0 < \gamma_1 < 1 < \gamma_2\).
    Set \(k = 0\), \(t = 2\), \(S_0 = 0\), and  Transient = True.
    \FOR{\(n=1,\ldots,N_{\text{epochs}}\)} 
    \IF{Transient is True}
    
    \STATE%
    Sample \(\xi_k\) and compute \(g(x_k, \xi_k)\).
    
    \STATE%
    Compute \(s_k = -\frac{1}{\sigma_k}g(x_k, \xi_k)\) and define \(x_{k+1} = x_k + s_k\).
    
    \STATE%
    Evaluate \(f(x_{k+1}, \xi_{k})\) and \(g(x_{k+1}, \xi_{k})\), and define
    \begin{equation}
      \bar{\rho}_k =  \frac{f(x_k, \xi_{k})-f(x_{k+1}, \xi_{k})}{\bar{T}_k(0)-\bar{T}_k(s_k)} = \frac{f(x_k, \xi_{k})-f(x_{k+1}, \xi_{k})}{\frac{1}{\sigma_k}\|g(x_k, \xi_{k})\|^2}.
    \end{equation}
    
    \STATE%
    Set
    \begin{equation}
      \label{ar:update3}
      \sigma_{k+1} \in 
      \begin{cases}
        [\max(\sigma_{\min}, \gamma_1\sigma_k), \sigma_k]
        &\text{ if } \bar{\rho}_k \geq \eta, \\
        [\sigma_k, \gamma_2\sigma_k]
        &\text{ if } \bar{\rho}_k < \eta. \\
      \end{cases}
    \end{equation}
    
    \STATE%
    Compute \(S_{k+1} = S_{k} + {g(x_{k+1}, \xi_{k})}^\top g(x_{k}, \xi_{k})\) and set \(m_{k+1} = m_{k}\).
    
    \STATE%
    If (\(k > \text{burn-in}\) and \(S_{k+1} < 0\)), then set Transient = False.
    
    \ELSE
    
    \STATE%
    Sample \(\xi_k\). Compute \(g(x_k, \xi_k)\) and \(\|V_{i \in \xi_k}(\nabla f_i(x_k))\|_1\).
    
    \IF{\eqref{eq:error-grad2_new} holds}
    
    \STATE%
    Go to step~\ref{step18:alg:regularization-inexact_new2}.
    
    \ELSE 
    
    \STATE%
    Set
    \begin{equation*}
        m_k = \min{\left( \ceil*{\frac{ \sigma_k^2 \|V_{i \in \xi_k}(\nabla f_i(x_k))\|_1}{\|g(x_k, \xi_k)\|^{2}}}, m_{\max}\right)}
    \end{equation*}
    \STATE%
    Sample a new \(\xi_k\) of new size \(m_k\) and compute \(g(x_k, \xi_k)\). 
    
    \ENDIF
    
    \STATE%
    \label{step18:alg:regularization-inexact_new2}
    Compute the step \(s_k = -\frac{1}{\sigma_k}g(x_k, \xi_k)\) and define \(x_{k+1} = x_k + s_k\).
    
    \STATE%
    Set \(m_{k+1} = m_{k}\), \(\sigma_{k+1} = \sigma_{k} \frac{t}{t-1}\), and increment t by one.
    
    \ENDIF
    \STATE%
    Increment \(k\) by one and go to step~\ref{step3:alg:regularization-inexact_new2}.
  \ENDFOR
  \end{algorithmic}
\end{algorithm}


\subsection{Experimental evaluation}%
\label{sec:p1-test}

We study the performance of ARAS in both convex and nonconvex settings. First, we compare it to SGD and SGD with momentum in the convex setting, where we consider a logistic regression problem. Then, we compare two versions of ARAS to SGD in the nonconvex setting, where we consider a nonconvex support vector machine problem with a sigmoid loss function. 

\subsubsection{Logistic regression with MNIST}%
\label{sec:log-mnist}

To evaluate the performance of our algorithm in the convex setting, we compare it to SGD and SGD with momentum for solving a logistic regression problem on the MNIST dataset. All parameters were set using a grid-search and all algorithms were trained for 10 epochs each. 

\begin{figure}[htbp]
  \centering
  \includegraphics[width=0.485\linewidth,trim=16 10 41 18]{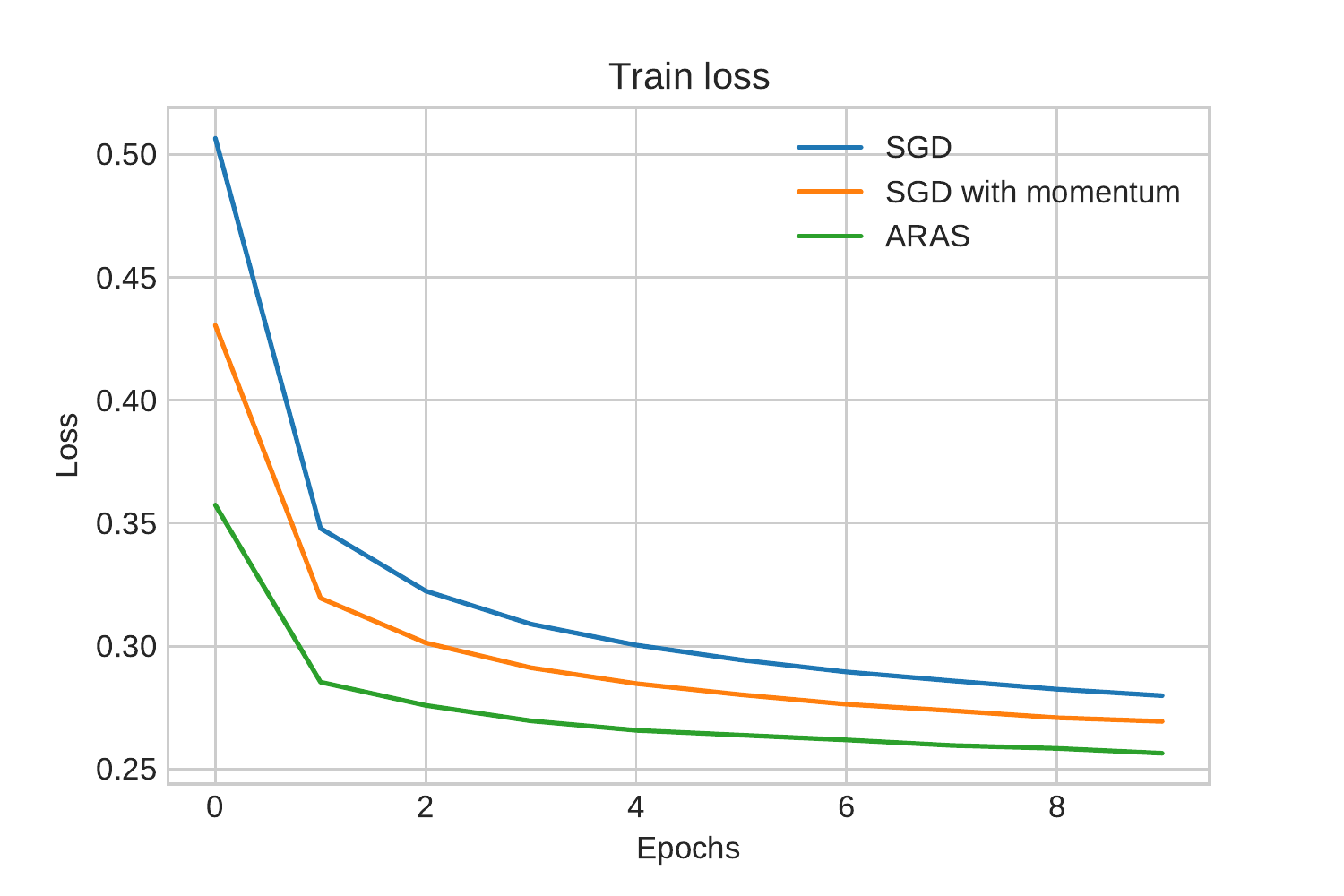}
  \hfill
  \includegraphics[width=0.485\linewidth,trim=16 10 41 18]{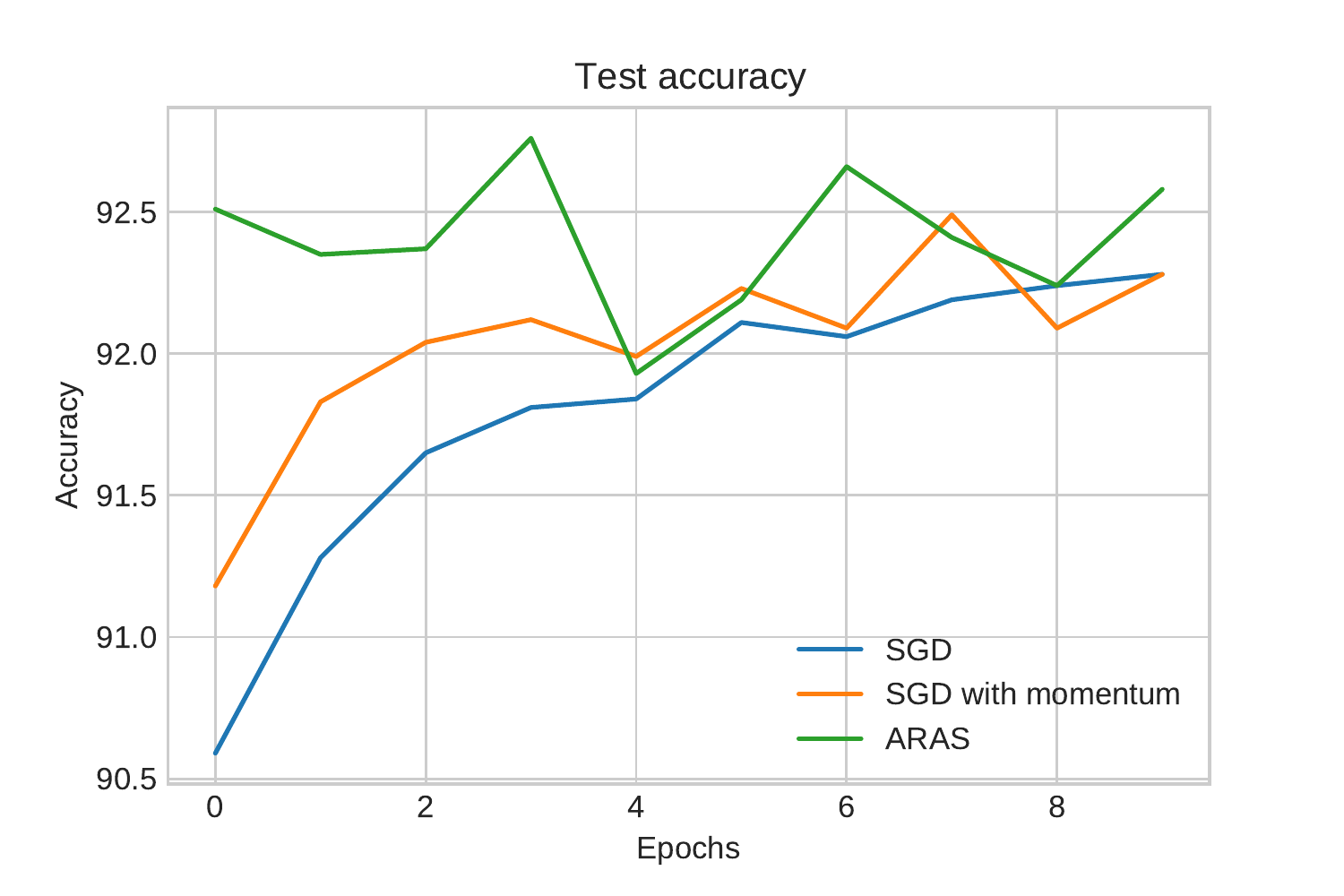}
  \caption{Evolution of the training loss (left) and the validation accuracy (right) for ARAS, SGD and SGD with momentum, solving a logistic regression problem on MNIST.\label{fig:MNIST_ARAS}}
\end{figure}

\Cref{fig:MNIST_ARAS} shows the performance of the three algorithms. ARAS outperforms SGD and SGD with momentum on both the training and test sets. 

\subsubsection{Nonconvex support vector machine with RCV1}%
\label{sec:svm-rcv1}

In the nonconvex setting, we compare the performance of ARAS and SGD for solving the finite-sum version of a nonconvex support vector machine problem, with a sigmoid loss function, a problem that was considered in \citet{wang2017stochastic}:

\begin{equation}
  \label{eq:svm}
  \min_{x \in \R^n} \mathbb{E}_{u,v}[1 - \tanh(v \, x^\top u)] + \lambda \|x\|^2,
\end{equation}
where \(u \in \R^n\) represents the feature vector, \(v \in \{-1,1\}\) represents the label value, and \(\lambda\) is the regularization parameter. 

The finite-sum version of~\eqref{eq:svm} can be written as
\begin{equation}
  \label{eq:svm1}
  \min_{x \in \R^n} \frac{1}{N} \sum_{i=1}^{N} f_i(x) + \lambda \|x\|^2,
\end{equation}
where \(f_i(x) = 1 - \tanh(v_i \, x^\top u_i)\) for \(i = 1,\ldots,N\). 

We use a subset\footnote{We downloaded the subset from http://www.cad.zju.edu.cn/home/dengcai/Data/TextData.html} of the dataset Reuters Corpus Volume I (RCV1) \citep{lewis2004rcv1}, which is a collection of manually categorized newswire stories. The subset contains \(9625\) stories with \(29992\) distinct words.
There are four categories that we reduce to the binary categories \(\{-1,1\}\)\footnote{The two categories ``MCAT'' and ``ECAT'' correspond to label value \(1\), and the two categories ``C15'' and ``GCAT'' correspond to label value \(-1\).}.
We then solve this binary classification problem.

\begin{figure}[htbp]
  \centering
    \includegraphics[width=0.485\linewidth,trim=16 10 41 18]{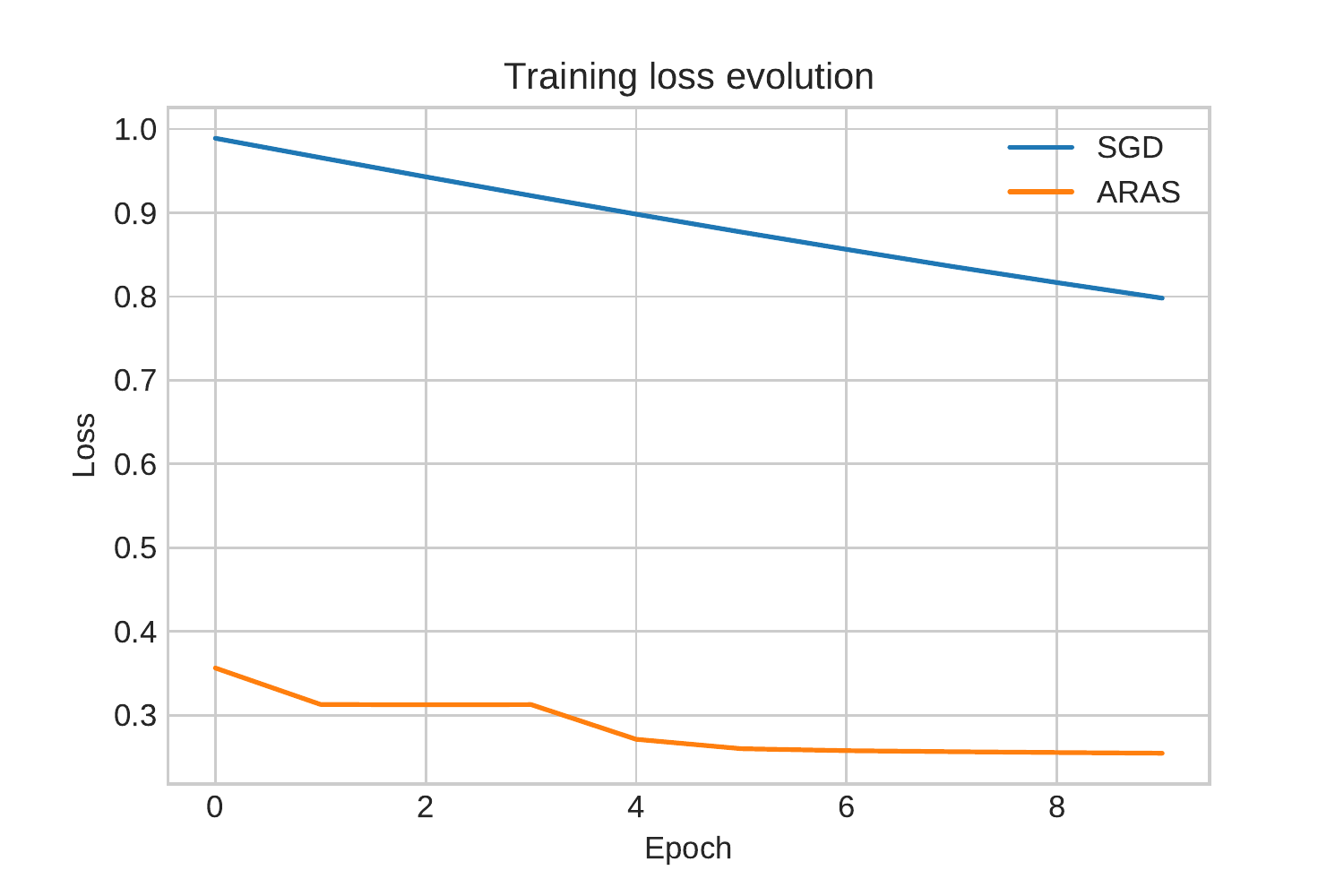}
    \hfill
    \includegraphics[width=0.485\linewidth,trim=16 10 41 18]{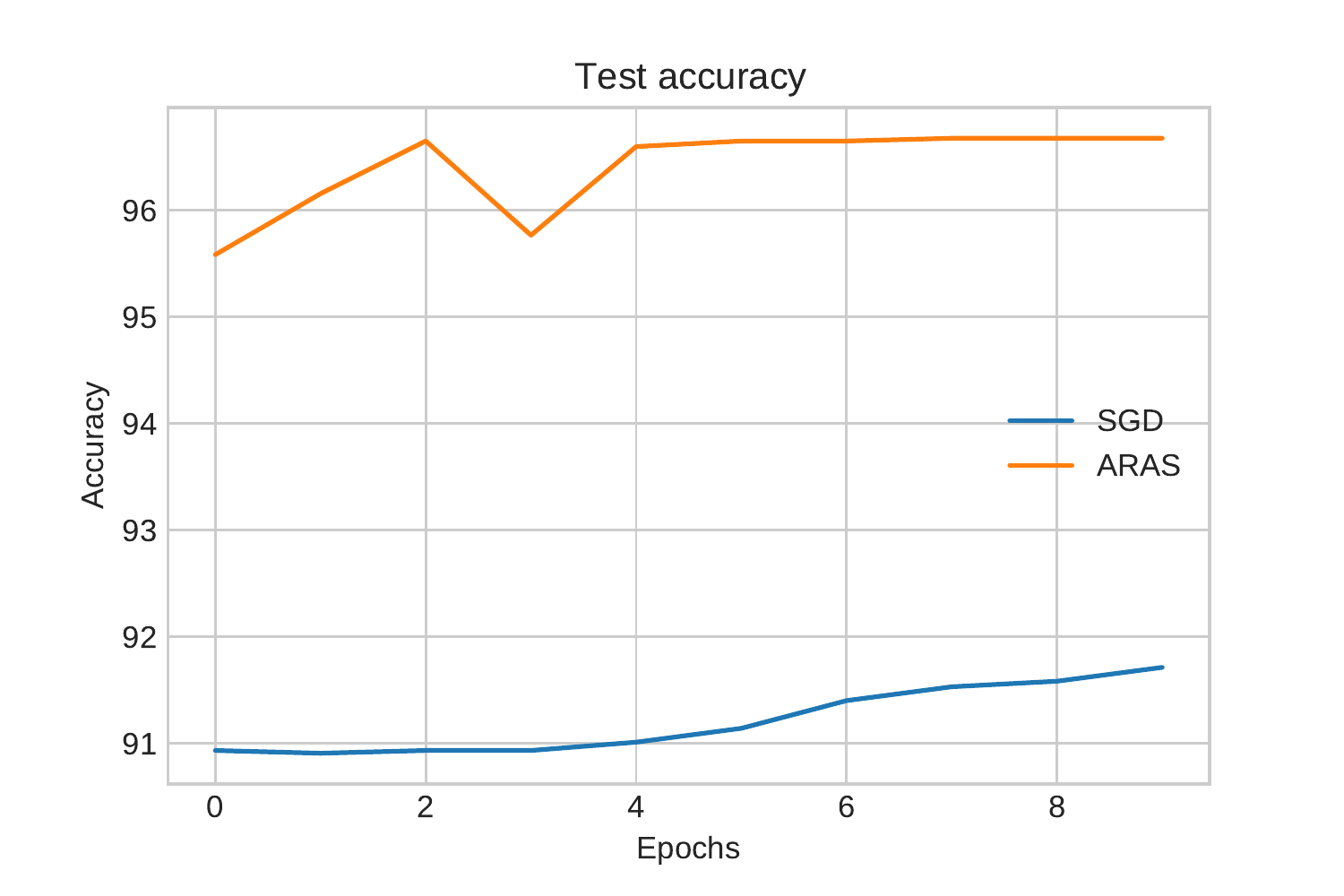}
    \caption{Evolution of the training loss (left) and the test accuracy (right) for SGD and ARAS solving a nonconvex SVM problem on RCV1.\label{fig:RCV1_ARAS}}
\end{figure}

\Cref{fig:RCV1_ARAS} reports our results; ARAS largely outperforms SGD both in terms of the training loss and test accuracy. 

\section{Stochastic damped L-BFGS with controlled norm of the Hessian approximation}%
\label{sec:secondorder}

In this section, we consider second-order methods and we assume that, at iteration \(k\), we can obtain a stochastic approximation~\eqref{eq:minibatch-gradient}
of \(\nabla f(x_k)\).
%
%
Iterates are updated according to~\eqref{eq:theta}, where this time \(B_k \neq I\).

The stochastic BFGS method \citep{schraudolph2007stochastic} computes an updated approximation \(H_{k+1}\) according to
\begin{equation}
  \label{eq:BFGS-update_4}
  H_{k+1} = V_k H_k V_k^\top + \rho_k s_k s_k^\top,
  \quad \text{where } \quad
  V_k = I - \rho_k s_k y_k^\top
  \quad \text{and} \quad
  \rho_k = 1 / s_k^\top  y_k,
\end{equation}
which ensures that the secant equation \(B_{k+1}s_k = y_k\) is satisfied, where
\begin{equation}
\label{eq:define-y-s}
  s_k = x_{k+1}-x_k, \quad \text{and} \quad
  y_k = g(x_{k+1}, \xi_k) - g(x_k, \xi_k).
\end{equation}
If \(H_k \succ 0\) and the curvature condition \(s_k^\top y_k > 0\) holds, then \(H_{k+1} \succ 0\) \citep{fletcher1970new}.

Because storing \(H_k\) and performing matrix-vector products is costly for large-scale problems, we use the limited-memory version of BFGS (L-BFGS) \citep{nocedal1980updating, liu1989limited}, in which \(H_k\) only depends on the most recent \(p\) iterations and an initial \(H_k^0 \succ 0\).
The parameter \(p\) is the \emph{memory}.
The inverse Hessian update can be written as
\begin{equation}
\label{eq:LBFGS-update_4}
\begin{aligned}
  H_k = &\ (V_{k-1}^\top  \dots V_{k-p}^\top ) H_k^0 (V_{k-p} \dots V_{k-1}) + \\
        & \rho_{k-p}(V_{k-1}^\top  \dots V_{k-p+1}^\top ) s_{k-p} s_{k-p}^\top  (V_{k-p+1} \dots V_{k-1}) + \cdots + 
        \rho_{k-1} s_{k-1}s_{k-1}^\top. 
\end{aligned}
\end{equation}
When \(\alpha_k\) in~\eqref{eq:theta} is not computed using a Wolfe line search \citep{wolfe1969convergence, wolfe1971convergence}, there is no guarantee that the curvature condition holds.
A common strategy is to simply skip the update.
By contrast, \citet{powell1978algorithms} proposed \emph{damping}, which consists in updating \(H_k\) using a modified \(y_k\), denoted by \(\hat{y}_k\), to benefit from information discovered at iteration \(k\) while ensuring sufficient positive definiteness.
We use
\begin{equation}
\label{eq:new-ytilde}
  \hat{y}_k := \theta_k y_k + (1-\theta_k) B_{k+1}^0 s_k,
\end{equation}
which is inspired by \citet{wang2017stochastic}, and differs from the original proposal of \citet{powell1978algorithms}, where
\begin{equation}
\label{eq:new_theta}
\theta_k =
1 \ \text{ if }\ s_k^\top y_k \geq \eta s_k^\top  B_{k+1}^0 s_k, \ \text{ and }\ (1-\eta)\frac{s_k^\top  B_{k+1}^0 s_k}{s_k^\top  B_{k+1}^0 s_k - s_k^\top  y_k} \text{ otherwise,}
\end{equation}
with \(\eta \in (0,1)\) and \(B_{k+1}^0 := {(H_{k+1}^0)}^{-1}\).
The choice~\eqref{eq:new-ytilde} ensures that the curvature condition
\begin{equation}
\label{eq:new-sTy-bounded}
    s_k^\top  \hat{y}_k \geq \eta s_k^\top  B_{k+1}^0 s_k \geq \eta \lambda_{\min}(B_{k+1}^0) \|s_k\|^2 > 0 ,
\end{equation}
is always satisfied since \(H_{k+1}^0 \succ 0\).
We obtain the damped L-BFGS update,
which is~\eqref{eq:LBFGS-update_4} with each \(V_i\) and \(\rho_i\) replaced with \(\hat{V}_i = I - \hat{\rho}_i s_i^\top \hat{y}_i\) and \(\hat{\rho}_i = 1 / s_i^\top  \hat{y}_i\).

In the next section, we propose a new version of stochastic damped L-BFGS that maintains estimates of the smallest and largest eigenvalues of \(H_k\). We show that the new algorithm requires less restrictive assumptions with respect to current literature, while being more robust to ill-conditioning and maintaining favorable convergence properties and complexity as established in \Cref{sec:convergence_4}. Finally, \Cref{sec:p2-test} provides a detailed experimental evaluation.

\subsection{A new stochastic damped L-BFGS with controlled Hessian norm}








Our working assumptions are \Cref{asm:f-C1,asm:g-lipschitz}
We begin by deriving bounds on the smallest and largest eigenvalues of \(H_{k+1}\) as functions of bounds on those of \(H_k\).
Proofs can be found in Appendix A.

\begin{lemma}%
  \label{lem:eigenvalues-update}
  Let \(s\) and \(y \in \mathbb{R}^n\) such that \(s^\top y \geq \gamma \|s\|^2\) with \(\gamma > 0\), and such that \(\|y\|\leq L_y \|s\|\), with \(L_y > 0\).
  Let \(A =\mu V V^\top  + \rho s s^\top \), where \(\rho = 1 / s^\top y\), \(\mu > 0\), and \(V = I - \rho s y^\top\).
  Then,
  \[
    0 <
    \min \left( \frac{1}{L_y}, \frac{\mu}{1 + \frac{\mu}{\gamma} L_y^2} \right) \leq
    \lambda_{\min}(A) \leq
    \lambda_{\max}(A) \leq
    \frac{1}{\gamma} + \max \left( 0, \frac{\mu}{\gamma^2}L_y^2 - \frac{\mu}{1 + \frac{\mu}{\gamma}L_y^2} \right).
  \]
\end{lemma}

In order to use \Cref{lem:eigenvalues-update} to obtain bounds on the eigenvalues of \(H_{k+1}\), we make the following assumption:
\begin{assumption}%
  \label{asm:stochastic-g-lipschitz}
  There is \(L_g > 0\) such that for all \(x\), \(y \in \R^n\), \(\|g(x, \xi) - g(y, \xi)\| \leq L_g \, \|x - y\|\).
\end{assumption}

\Cref{asm:stochastic-g-lipschitz} is required to prove convergence and convergence rates for most recent stochastic quasi-Newton methods \citep{yousefian2017smoothing}.
It is less restrictive than requiring \(f(x,\xi)\) to be twice differentiable with respect to \(x\), and the Hessian \(\nabla_{xx}^{2}f(x,\xi)\) to exist and be bounded for all \(x\) and \(\xi\), as in \citet{wang2017stochastic}. 

The next theorem shows that the eigenvalues of \(H_{k+1}\) are bounded and bounded away from zero.

\begin{theorem}%
  \label{thm:new-L-BFGS-update}
  Let \Cref{asm:f-C1,asm:g-lipschitz,asm:stochastic-g-lipschitz} hold. Let \(H_{k+1}^0 \succ 0\) and \(p > 0\).
  If \(H_{k+1}\) is obtained by applying \(p\) times the damped BFGS update formula with inexact gradient to \(H_{k+1}^0\), there exist easily computable constants \(\lambda_{k+1}\) and \(\Lambda_{k+1}\) that depend on \(L_g\) and \(H_{k+1}^0\) such that \(0 < \lambda_{k+1} \leq \lambda_{\min}(H_{k+1}) \leq \lambda_{\max}(H_{k+1})\leq \Lambda_{k+1}\).

\end{theorem}

The precise form of \(\lambda_{k+1}\) and \(\Lambda_{k+1}\) is given in~\eqref{eq:lambdaminpr} and~\eqref{eq:lambdamaxpr2} in \Cref{sec-appendix}.



A common choice for \(H_{k+1}^0\) is
\(
    H_{k+1}^0 = \gamma_{k+1}^{-1} I 
\)
where \(\gamma_{k+1} = y_{k}^\top y_k / s_k^\top y_k\),
is the \emph{scaling parameter}. This choice ensures that the search direction is well scaled, which promotes large steps.
To keep \(H_{k+1}^0\) from becoming nearly singular or non positive definite, 
we define
\begin{equation}
\label{eq:init_HK}
    H_{k+1}^0 = \left( \max(\underline{\gamma}_{k+1}, \min(\gamma_{k+1},\overline{\gamma}_{k+1})) \right) I, 
\end{equation}
where \(0 < \underline{\gamma}_{k+1} < \overline{\gamma}_{k+1}\) can be constants or iteration dependent.



The Hessian-gradient product used to compute \(d_k = - H_k g(x_k,\xi_k)\) can be obtained cheaply by exploiting a recursive algorithm \citep{nocedal1980updating, lotfi2021stochastic}.

Motivated by the success of recent methods combining variance reduction with stochastic L-BFGS \citep{gower2016stochastic, moritz2016linearly, wang2017stochastic}, we apply an SVRG-like type of variance reduction \citep{svrg} to the update.
Not only does this accelerate the convergence, since we can choose a constant step size, but it also improves the quality of the curvature approximation.

We summarize our complete algorithm,  VAriance-Reduced stochastic damped L-BFGS with Controlled HEssian Norm (VARCHEN), as \Cref{alg:VR-SdLBFGS-CHN}.

\begin{algorithm}[htbp]%
  \label{alg:VR-SdLBFGS-CHN}
  \caption{Variance-Reduced Stochastic Damped L-BFGS with Controlled Hessian Norm}
  \begin{algorithmic}[1]

    \STATE%
    Choose \(x_0 \in \R^n\), step size sequence \(\{\alpha_k > 0\}_{k\geq 0}\), batch size sequence \(\{m_k > 0\}_{k\geq 0}\), eigenvalue limits \(\lambda_{\max} > \lambda_{\min} > 0\), memory parameter \(p\), total number of epochs \(N_{\text{epochs}}\), and sequences \(\{\underline{\gamma}_{k} > 0\}_{k\geq 0}\) and \(\{\overline{\gamma}_{k+1} > 0\}_{k\geq 0}\), such that
    \(0 < \lambda_{\min} < \underline{\gamma}_{k} < \overline{\gamma}_{k} < \lambda_{\max}\), for every \(k \geq 0\). Set \(k = 0\) and \(H_0 = I\).
    
    \FOR{\(t=1,\ldots,N_{\text{epochs}}\)}

      \STATE%
      \label{step3:alg:VR-SdLBFGS-CHN}
      Define \(x_k^t = x_k\) and compute the full gradient \(\nabla f(x_k^t)\).
      Set \(M = 0\).

      \WHILE{\(M < N\)}
      
        \STATE%
        \label{step4:alg:VR-SdLBFGS-CHN}
        Sample batch \(\xi_k\) of size \(m_k \leq N - M\) and compute \(g(x_k,\xi_k)\) and \(g(x_k^t,\xi_k)\).
        
        \STATE%
        \label{step5:alg:VR-SdLBFGS-CHN}
        Define \(\Tilde{g}(x_k,\xi_k) = g(x_k,\xi_k) - g(x_k^t,\xi_k) + \nabla f(x_k^t)\).
        
        \STATE%
        \label{step9:alg:VR-SdLBFGS-CHN}
        Estimate \(\Lambda_k\) and \(\lambda_k\) in \Cref{thm:new-L-BFGS-update}. 
        If \(\Lambda_k > \lambda_{\max}\) or \(\lambda_k < \lambda_{\min}\), delete  \(s_i\), \(y_i\) and \(\hat{y}_i\) for \(i = k-p+1,\ldots,k-2\). 
        
        \STATE%
        Compute \(d_k = - H_k \Tilde{g}(x_k,\xi_k)\).
        
        \STATE%
        Define \(x_{k+1} = x_k + \alpha_k d_k\), and compute \(s_k\), \(y_k\) as in~\eqref{eq:define-y-s}, and \(\hat{y}_k\) as in~\eqref{eq:new-ytilde}.
        
        \STATE%
        Increment \(k\) by one and update \(M \leftarrow M + m_k\).
      
      \ENDWHILE

    \ENDFOR
  \end{algorithmic}
\end{algorithm}

In step~\ref{step9:alg:VR-SdLBFGS-CHN} of \Cref{alg:VR-SdLBFGS-CHN}, we compute an estimate of the upper and lower bounds on \(\lambda_{\max}(H_{k})\) and \(\lambda_{\min}(H_{k})\), respectively. The only unknown quantity in the expressions of \(\Lambda_k\) and \(\lambda_k\) in \Cref{thm:new-L-BFGS-update} is \(L_g\), which we estimate as \(L_g \approx L_{g,k} := \|y_k\| / \|s_k\|\).
When the estimates are not within the limits \([\lambda_{\min}, \, \lambda_{\max}]\),  
we delete \(s_i\), \(y_i\) and \(\hat{y}_i\), \(i \in \{k-p+1,\ldots,k-2\}\) from storage, such that \(H_{k} g(x_{k},\xi_{k})\) is computed using the most recent pair \((s_{k-1}, \hat{y}_{k-1})\) only and \(d_{k} = - H_{k} g(x_{k},\xi_{k})\).
Although it is not theoretically guaranteed that the eigenvalues of the new Hessian approximation using the most recent pain are within the bounds, we found that it is the case in practice and that this strategy yields better results than setting the approximation to the identity.
Finally, a full gradient is computed once in every epoch in step~\ref{step3:alg:VR-SdLBFGS-CHN}.
The term \(g(x_k^t,\xi_k) - \nabla f(x_k^t)\) can be seen as the bias in the gradient estimation \( g(x_k,\xi_k)\), and it is used here to correct the gradient approximation in step~\ref{step5:alg:VR-SdLBFGS-CHN}.

\subsection{Convergence and complexity analysis}%
\label{sec:convergence_4}

We show that \Cref{alg:VR-SdLBFGS-CHN} satisfies the assumptions of the convergence analysis and iteration complexity of \citet{wang2017stochastic} for stochastic quasi-Newton methods.
We make an additional assumption used by \citet{wang2017stochastic} to establish global convergence.

\begin{assumption}%
  \label{asm:bounded-exp}
  For all \(k\), \(\xi_k\) is independent of \(\{x_1, \ldots, x_k\}\), \(\E_{\xi_k}\left[ g(x_k, \xi_k) \right] = \nabla f(x_k)\), and there exists \(\sigma > 0\) such that \(\E_{\xi_k}[ \| g(x_k, \xi_k) - \nabla f(x_k) \|^2 ] \leq \sigma^2\).
\end{assumption}

Our first result follows from \citet[Theorem~\(2.6\)]{wang2017stochastic}, whose remaining assumptions are satisfied as a consequence of~\eqref{eq:minibatch-gradient}, \Cref{thm:new-L-BFGS-update}, the mechanism of \Cref{alg:VR-SdLBFGS-CHN},~\eqref{eq:BFGS-update_4} and our choice of \(\alpha_k\) below.

\begin{theorem}
  \label{lem:convergence-theorem}
  Assume \(m_k = m\) for all \(k\), that \Cref{asm:f-C1,asm:g-lipschitz,asm:stochastic-g-lipschitz,asm:bounded-exp} hold for \(\{x_k\}\) generated by \Cref{alg:VR-SdLBFGS-CHN}, and that \(\alpha_k := c / (k+1)\) where \(0 < c \leq \lambda_{\min} / (L \lambda_{\max})\).
  Then, \(\liminf \|\nabla f(x_k)\| = 0\) with probability \(1\).
  Moreover, there is \(M_f > 0\) such that \(\E[f(x_k)] \leq M_f\) for all \(k\). If we additionally assume that there exists \(M_g > 0\) such that \(\E_{\xi_k}[\|g(x_k, \xi_k)\|^2] \leq M_g\), then  \(\lim \|\nabla f(x_k)\| = 0\) with probability \(1\).
\end{theorem}

Our next result follows in the same way from \citet[Theorem~\(2.8\)]{wang2017stochastic}.

\begin{theorem}
  \label{lem:convergence-theorem2}
  Let assumptions of \Cref{lem:convergence-theorem} be satisfied.
  Assume in addition that there exists \(f_{\text{low}}\) such that \(f(x_k) \geq f_{\text{low}}\) for all \(k\).
  Let \(\alpha_k := \lambda_{\min} / (L \lambda_{\max}^2) k^{-\beta}\) for all \(k > 0\), with \(\beta \in (\tfrac{1}{2}, \, 1)\).
  Then, for all \(T > 0\),
    \begin{equation}
      \frac{1}{T} \sum_{k=1}^{T} \E\left[ \| \nabla f(x_k)\|^2 \right] \leq
      \frac{2 L (M_f - f_{\text{low}}) \Lambda^2 }{\lambda^2} T^{\beta - 1} + \frac{\sigma^2}{(1 - \beta) m} (T^{-\beta} - T^{-1}).
    \end{equation}
    Moreover, for any \(\epsilon \in (0, \, 1)\), we achieve
    \begin{equation*}
        \frac{1}{T} \sum_{k=1}^{T} \E\left[ \| \nabla f(x_k)\|^2 \right] \leq \epsilon.
    \end{equation*}
    after at most \(T = \BigO(\epsilon^{-1 / (1 - \beta)})\) iterations.
\end{theorem}

\subsection{Experimental evaluation}%
\label{sec:p2-test}

We compare VARCHEN to SdLBFGS-VR \citep{wang2017stochastic} and to SVRG \citep{svrg} for solving a multi-class classification problem.
We train a modified version\footnote{\href{https://colab.research.google.com/github/pytorch/ignite/blob/master/examples/notebooks/Cifar10_Ax_hyperparam_tuning.ipynb}{FastResNet Hyperparameters tuning with Ax on CIFAR10}} of the deep neural network model DavidNet\footnote{\href{https://myrtle.ai/learn/how-to-train-your-resnet-4-architecture/}{https://myrtle.ai/learn/how-to-train-your-resnet-4-architecture/}} proposed by David C. Page, on CIFAR-10 \citep{Krizhevsky09learningmultiple} for \(20\) epochs. 
Note that we also used VARCHEN and SdLBFGS-VR  to solve a logistic regression problem using the MNIST dataset \citep{lecun2010mnist} and a nonconvex support-vector machine problem with a sigmoid loss function using the RCV1 dataset \citep{lewis2004rcv1}.
The performance of both algorithms are on par on those problems because, in contrast with DavidNet on CIFAR-10, they are not highly nonconvex or ill conditioned.

\begin{figure}[t]
  \centering
    \includegraphics[width=.485\linewidth,trim=16 10 41 18]{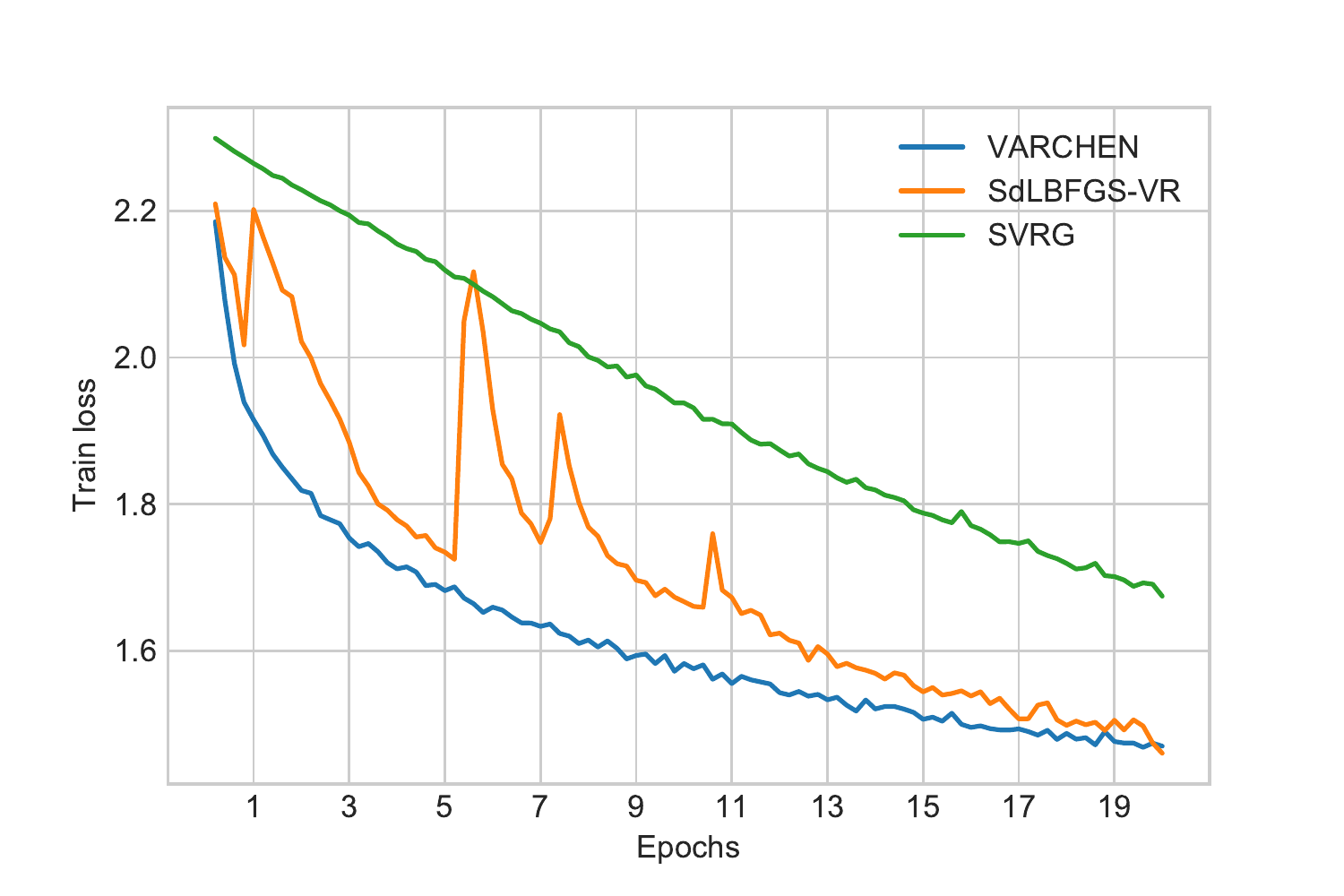}
    \hfill
    \includegraphics[width=.485\linewidth,trim=16 10 41 18]{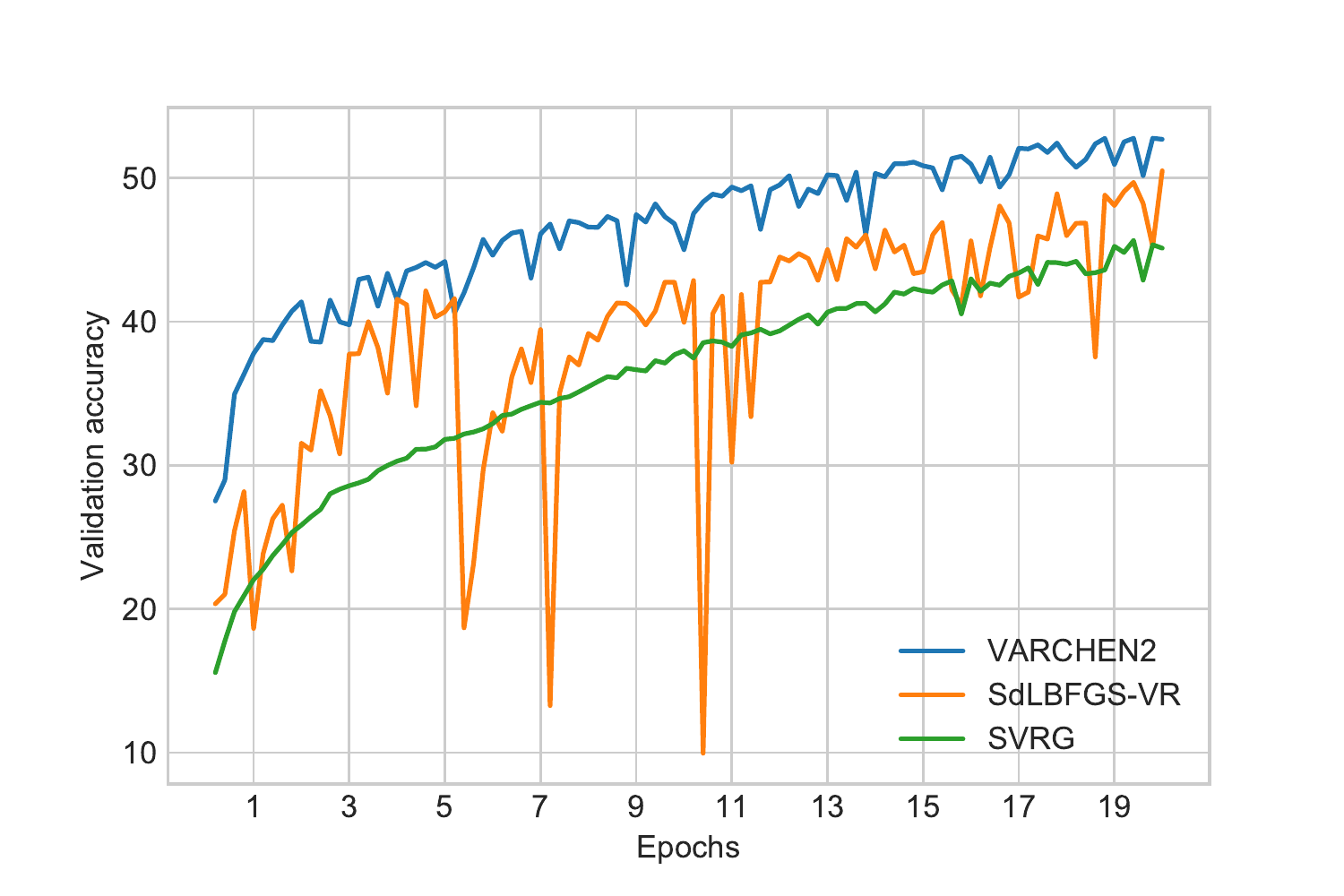}
    \caption{Evolution of the training loss (left) and the validation accuracy (right) for training a modified DavidNet on CIFAR-10.\label{fig:CIFAR-Davnet1}}
\end{figure}

\begin{figure}[t]
  \centering
    \includegraphics[width=.485\linewidth,trim=16 10 41 18]{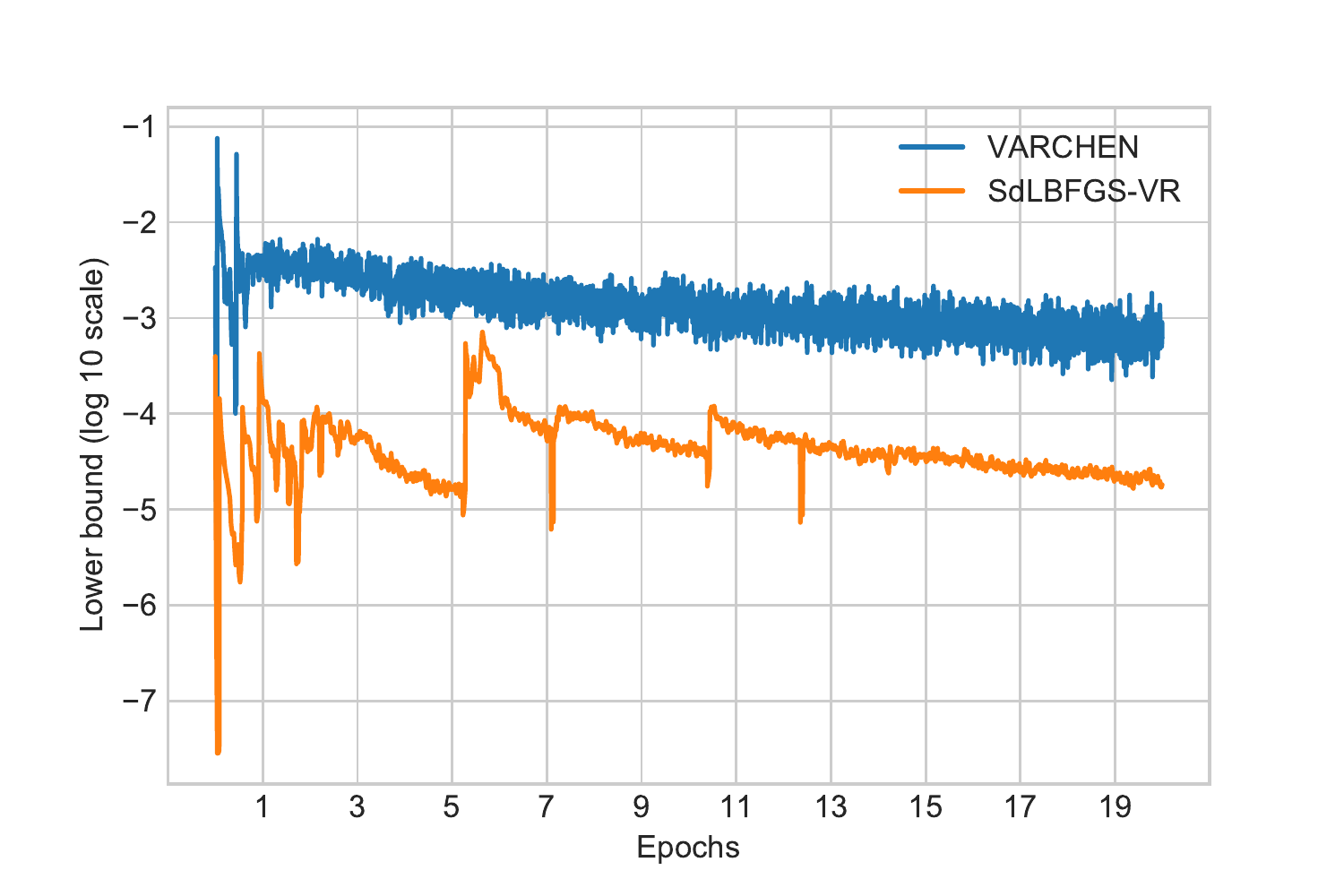}
    \hfill
    \includegraphics[width=.485\linewidth,trim=16 10 41 18]{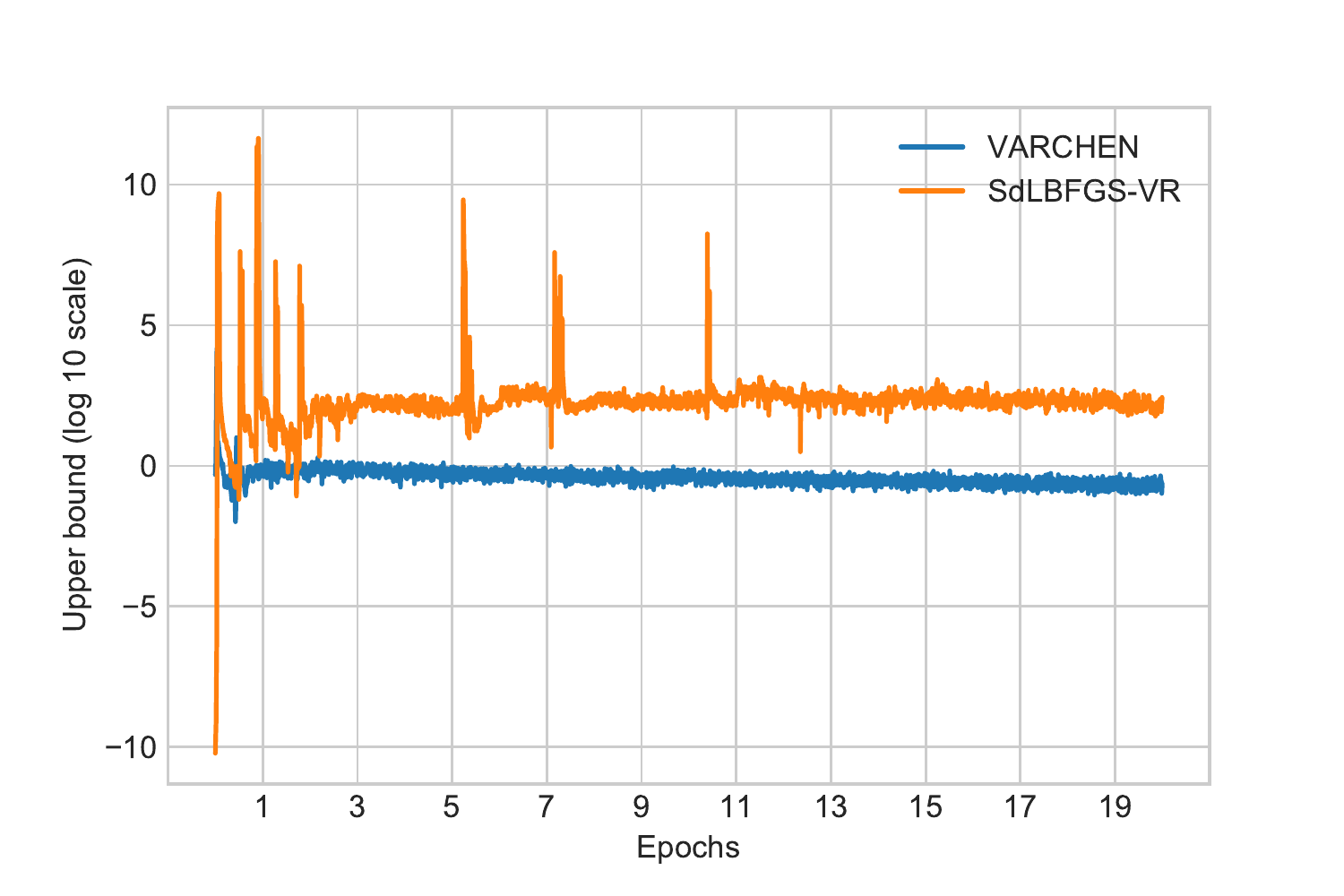}
    \caption{Evolution of the lower bound on the smallest eigenvalue \(\lambda_k\) (left) and the upper bound on the largest eigenvalue \(\Lambda_k\) (right) on a base 10 logarithmic scale for training a modified DavidNet on CIFAR-10.\label{fig:CIFAR-Davnet2}}
\end{figure}

\Cref{fig:CIFAR-Davnet1} shows that VARCHEN outperforms SdLBFGS-VR for the training loss minimization task, and both outperform SVRG\@.
VARCHEN has an edge over SdLBFGS-VR in terms of the validation accuracy, and both outperform SVRG\@.
More importantly, the performance of VARCHEN is more consistent than that of SdLBFGS-VR, displaying a smoother, less oscillatory behaviour.
To further investigate this observation, we plot the evolution of \(\Lambda_k\) and \(\lambda_k\) as shown in \Cref{fig:CIFAR-Davnet2}. We see that the estimate of the lower bound on the smallest eigenvalue is smaller for SdLBFGS-VR compared to VARCHEN\@. 
We also notice that the estimate of the upper bound of the largest eigenvalue of \(H_k\) takes even more extreme values for SdLBFGS-VR compared to VARCHEN\@. 
The extreme values \(\lambda_k\) and \(\Lambda_k\) reflect an ill-conditioning problem encountered when using SdLBFGS-VR and we believe that it explains the extreme oscillations in the performance of SdLBFGS-VR\@.

\section{Conclusions and future work}%
\label{sec:final}

In this paper, we adapt deterministic first- and second-order methods for large-scale optimization to the stochastic context of machine learning, while preserving convergence guarantees and iteration complexity.

For first-order methods, we propose an adaptation of ARIG to the stochastic setting that maintains the same theoretical guarantees. 
However, this new version is still not practical for solving machine learning problems since it requires evaluating the full gradient to test the accuracy of the gradient approximation. 
Additionally, objective approximations are required to be increasingly accurate as convergence occurs. 
Therefore, we propose to leverage adaptive sampling to satisfy the accuracy condition on the gradient approximation.
\citet{lotfi2020stochastic} shows that the resulting algorithm compares favorably with SGD while maintaining batch size reasonable. 
Finally, we propose ARAS, which addresses both the practicality and the overall convergence concerns. The computational experiments show that ARAS outperforms SGD and enjoys a conceivable overall convergence property. Nevertheless, its convergence rate and iteration complexity remain an open question.

Concerning second-order methods, we use the stochastic damped L-BFGS algorithm in a nonconvex setting, where there are no guarantees that \(H_k\) remains well-conditioned and numerically nonsingular throughout.
We introduce a new stochastic damped L-BFGS algorithm that monitors the quality of \(H_k\) during the optimization by maintaining bounds on its largest and smallest eigenvalues.
Our work is the first to address the Hessian singularity problem by approximating and leveraging such bounds.
Moreover, we propose a new initial inverse Hessian approximation that results in a smoother, less oscillatory training loss and validation accuracy evolution.
Additionally, we use variance reduction in order to improve the quality of the curvature approximation and accelerate convergence.
Our algorithm converges almost-surely to a stationary point and numerical experiments show that it is more robust to ill-conditioning and more suitable to the highly nonconvex context of deep learning than SdLBFGS-VR\@. We consider this work to be a first step towards the use of bound estimates to control the quality of the Hessian approximation in approximate second-order algorithms. 
Future work should aim to improve the quality of these bounds and explore another form of variance reduction that consists of adaptive sampling \citep{jalilzadeh2018variable,bollapragada2018progressive,bollapragada2019adaptive}.

The first- and second-order algorithms that we propose in this paper fall under the umbrella of adaptive methods that are crucial today to reduce the alarming engineering energy and computational costs of hyperpatameter tuning of deep neural networks. 
Such algorithms will not only have a profound impact on the efficiency of machine learning methods, but they will also allow entities with limited computational resources to benefit from these methods.

\small
\bibliographystyle{abbrvnat}
\bibliography{abbrv,bib}
\normalsize

\clearpage
\appendix

\makeatletter
\def\@seccntformat#1{\csname Pref@#1\endcsname \csname the#1\endcsname\quad}
\def\Pref@section{Appendix~}

\def\@Opargbegintheorem#1#2#3#4{#4\trivlist
      \item[\hskip\labelsep{#3#1}]{#3#2\@thmcounterend\ }}
\makeatother

\section{Proofs}%
\label{sec-appendix}

\begin{proof}[of \Cref{lem:sigma-bounded}]
  By definition of \(L\), Taylor's theorem yields \(|f(x + s) - T(x, s)| \leq \tfrac{1}{2} L \|s\|^2\) for all \(x\) and \(s\).
  We deduce from~\eqref{eq:g-error} that \(|T(x_k, s_k) - \bar{T}_k(s_k)| \leq \omega_g^k \|\nabla f(x_k,\omega_g^k)\| \|s_k\|\) for all \(k\).
  Together, those two observations, the triangle inequality, the definition of \(s_k\) in Step~\ref{step3:alg:regularization-inexact} of \Cref{alg:regularization-inexact} and the bound \(\omega_k^g \leq \sigma_k^{-1}\) yield
  \[
    |\rho_k - 1| =
    \frac{|f(x_k + s_k) - \bar{T}_k(s_k)|}{\sigma_k^{-1} \|\nabla f(x_k,\omega_g^k)\|^2} \leq
    \frac{\tfrac{1}{2} L \sigma_k^{-2} + \omega_k^g \sigma_k^{-1}}{\sigma_k^{-1}} \leq
    (\tfrac{1}{2} L + 1) \sigma_k^{-1}.
  \]
  Thus, \(\sigma_k \geq (\tfrac{1}{2} L + 1) / (1 - \eta_2)\) \(\Longrightarrow\) \(\rho_k \geq \eta_2\), which means that iteration \(k\) is very successful and \(\sigma_{k+1} \leq \sigma_k\).
  In view of~\eqref{ar:update}, \(\sigma_k\) can never exceed the previous threshold by a factor larger than \(\gamma_3\), except if \(\sigma_0\) were larger than the latter bound.
  Overall, we obtain~\eqref{eq:sigma-max}. \qed
\end{proof}

\begin{proof}[\Cref{lem:inexact-decrease}]
Suppose two values of the error threshold \(\omega_f^k\) and \(\hat{\omega}_f^k\), such that, 

\begin{align}
  \label{eq:error1}
  |f(x_k, \omega_f^k) - f(x_k)| & \leq  \omega_f^k, \\
  \label{eq:error2}
  |f(x_k+s_k, \hat{\omega}_f^k) - f(x_k + s_k)| & \leq \hat{\omega}_f^k .
\end{align}

We also suppose a constant \(\eta_0 < \frac{1}{2} \eta_1\) and we require that 

\begin{equation}
\label{eq:important-condition}
    \max(\omega_f^k, \hat{\omega}_f^k) \leq \eta_0  \left[\bar{T}_k(0)-\bar{T}_k(s_k)\right]. 
\end{equation}

Using~\eqref{eq:error1} and~\eqref{eq:important-condition}, we deduce that

\begin{equation}
\label{eq:sto-f-ine1}
     f(x_k, \omega_f^k) - f(x_k) \leq  |f(x_k, \omega_f^k) - f(x_k)| \leq \omega_f^k \leq \eta_0 \left[\bar{T}_k(0)-\bar{T}_k(s_k)\right].
\end{equation}

Using~\eqref{eq:error2} and~\eqref{eq:important-condition}, we have that 

\begin{align}
\label{eq:sto-f-ine2}
     f(x_k+s_k, \hat{\omega}_f^k) - f(x_k + s_k) & \leq  |f(x_k+s_k, \hat{\omega}_f^k) - f(x_k +s_k)|, \\ & \leq \hat{\omega}_f^k
      \leq \eta_0  [\bar{T}_k(0)-\bar{T}_k(s_k)].
\end{align}

From~\eqref{eq:sto-f-ine1} and~\eqref{eq:sto-f-ine2}, we have that 

\begin{align}
\label{eq:sto-f-ine3}
\beta_k := \frac{\left[f(x_k, \omega_f^k) - f(x_k)\right] - \left[f(x_k+s_k, \hat{\omega}_f^k) - f(x_k + s_k)\right]}{\bar{T}_k(0)-\bar{T}_k(s_k)} \, \leq \,  \frac{\omega_f^k + \hat{\omega}_f^k}{\bar{T}_k(0)-\bar{T}_k(s_k)} \, \leq \, 2 \eta_0.
\end{align}

Notice that in this case, we have 

\begin{equation}
\label{eq:rho-extended}
\bar{\rho}_k := \frac{f(x_k, \omega_f^k) - f(x_k+s_k, \hat{\omega}_f^k)}{\bar{T}_k(0)-\bar{T}_k(s_k)} = \frac{f(x_k)-f(x_k+s_k)}{\bar{T}_k(0)-\bar{T}_k(s_k)} + \beta_k = \rho_k + \beta_k .
\end{equation}

Combining~\eqref{eq:sto-f-ine3} and~\eqref{eq:rho-extended}, we obtain the following implication:

\begin{equation}
    \bar{\rho}_k \geq \eta_1 \implies \rho_k \geq \eta_1 - 2 \eta_0 \geq \bar{\eta}_1 > 0,
\end{equation}

where \(\bar{\eta}_1\) is by definition equal to \(\eta_1 - 2 \eta_0\). 
\qed
\end{proof}

\begin{proof}[of \Cref{lem:eigenvalues-update}]
First notice that
\begin{equation*}
    \gamma\|s\|^2 \leq s^\top y \leq \|s\|\|y\|,
    \quad \text{and thus} \quad \|s\| \leq \frac{1}{\gamma}\|y\|.
\end{equation*}

Therefore, 

\begin{equation}
    \label{eq:rho-bounded}
    \frac{1}{\|s\|\|y\|} \leq \rho \leq \frac{1}{\gamma}\frac{1}{\|s\|^2}.
\end{equation}

  Since \(A\) is a real symmetric matrix, the spectral theorem states that its eigenvalues are real and it can be diagonalized by an orthogonal matrix.
  That means that we can find \(n\) orthogonal eigenvectors and \(n\) eigenvalues counted with multiplicity.

  Consider first the special case where \(s\) and \(y\) are collinear, i.e., there exists \(\theta > 0\) such that \(y = \theta s\).
  Any vector such that \(u \in s^\perp\), where \(s^\perp = \{x \in \R^n : x^\top s = 0\}\), is an eigenvector of \(A\) associated with the eigenvalue \(\mu\) of multiplicity \(n-1\).
  Moreover, \(s^\top y = \theta \|s\|^2=\|s\|\|y\|\), \(\rho = 1 / (\theta \|s\|^2)\) and we have
  \begin{equation*}
      As   = \left[\mu {\left(I-\rho\theta ss^\top \right)}^2 + \rho ss^\top \right] s
           = \left[\mu {\left( 1-\rho\theta\|s\|^2 \right)}^2 + \rho\|s\|^2 \right] s
           = \rho \|s\|^2 s.
  \end{equation*}
  Let us call \(\lambda = \rho\|s\|^2\), the eigenvalue associated with eigenvector \(s\).
  From~\eqref{eq:rho-bounded} and  \(\|y\|\leq L_y \|s\|\), we deduce that
  \begin{equation*}
      \frac{1}{L_y} \leq \lambda \leq \frac{1}{\gamma}.
  \end{equation*}

  Suppose now that \(s\) and \(y\) are linearly independent.
  Any \(u\) such that \(u^\top s = 0 = u^\top y\) satisfies \(Au = \mu u\).
  This provides us with a (\(n-2\))-dimensional eigenspace \(S\), associated to the eigenvalue \(\mu\) of multiplicity \(n-2\).
  Note that
  \begin{align*}
      As &= \rho \|s\|^2\ (1 + \mu \rho \|y\|^2)s - \rho \|s\|^2 \mu y, \\
      Ay &= s.
  \end{align*}
  Thus neither \(s\) nor \(y\) is an eigenvector of \(A\).
  Now consider an eigenvalue \(\lambda\) associated with an eigenvector \(u\), such that \(u \in S^\perp\).
  Since \(s\) and \(y\) are linearly-independent, we can search for \(u\) of the form \(u = s + \beta y\) with \(\beta > 0\).
  The condition \(A u = \lambda u\) yields
  \begin{align*}
    \rho \|s\|^2\ (1 + \mu \rho \|y\|^2) + \beta &= \lambda, \\
    -\rho \|s\|^2\mu &= \lambda \beta.
  \end{align*}
  We eliminate \(\beta = \lambda - \rho \|s\|^2\ (1 + \mu \rho \|y\|^2)\) and obtain
  \begin{equation*}
    p(\lambda) = 0,
  \end{equation*}
  where
  \begin{equation*}
      p(\lambda) = \lambda^2 - \lambda \rho \|s\|^2\ (1 + \mu \rho \|y\|^2) + \rho \|s\|^2\mu .
  \end{equation*}
  The roots of \(p\) must be the two remaining eigenvalues \(\lambda_1 \leq \lambda_2\) that we are looking for.
  In order to establish the lower bound, we need a lower bound on \(\lambda_1\) whereas to establish the upper bound, we need an upper bound on \(\lambda_2\).

  On the one hand, let \(l\) be the tangent to the graph of \(p\) at \(\lambda = 0\), defined by
  \begin{equation*}
      l(\lambda) = p(0) + p'(0)\lambda = \mu \rho \|s\|^2 - \lambda \rho \|s\|^2\left(1 + \mu \rho \|y\|^2\right).
  \end{equation*}
  Its unique root is
  \begin{equation*}
      \bar{\lambda} = \frac{\mu}{1 + \mu \rho \|y\|^2}.
  \end{equation*}
  From~\eqref{eq:rho-bounded} and since \(\|y\|\leq L_y \|s\|\), we deduce that
  \begin{equation*}
  \label{eq:lambda-bar-bounded}
    \bar{\lambda} \geq \frac{\mu}{1 + \frac{\mu}{\gamma} \frac{\|y\|^2}{\|s\|^2}} \geq \frac{\mu}{1+ \frac{\mu}{\gamma} L_y^2}.
  \end{equation*}
  Since \(p\) is convex, it remains above its tangent, and \(\bar{\lambda} \leq \lambda_1\).
  
  Finally, 
  \[
    \lambda_{\min}(A) \geq \min \left( \frac{1}{L_y}, \frac{\mu}{1+ \frac{\mu}{\gamma} L_y^2} \right) > 0.
  \]
  
  This establishes the lower bound.

  On the other hand, the discriminant \(\Delta = \rho^2 \|s\|^4 {(1 + \mu \rho \|y\|^2)}^2 - 4 \rho \|s\|^2 \mu\) must be nonnegative since \(A\) is real symmetric, and its eigenvalues are real.
  We have
  \begin{equation*}
      \lambda_2 = \frac{ \rho \|s\|^2 (1 + \mu \rho \|y\|^2) + \sqrt{ \rho^2 \|s\|^4 {(1 + \mu \rho \|y\|^2)}^2 - 4 \rho \|s\|^2 \mu } }{ 2 }.
  \end{equation*}
  For any positive \(a\) and \(b\) such that \(a^2 - b > 0\), we have \(\sqrt{a^2-b} \leq a - \tfrac{b}{2a}\). Thus,
  \begin{equation*}
      \lambda_2 \leq \rho \|s\|^2 (1 + \mu\rho\|y\|^2) - \frac{\mu}{1+\mu\rho\|y\|^2}.
  \end{equation*}
  From~\eqref{eq:rho-bounded}, we deduce that
  \begin{equation*}
      \lambda_2 \leq \frac{1}{\gamma} + \frac{\mu}{\gamma^2}\frac{\|y\|^2}{\|s\|^2} - \frac{\mu}{1 + \frac{\mu}{\gamma}\frac{\|y\|^2}{\|s\|^2}}.
  \end{equation*}
  And since \(\|y\|\leq L_y \|s\|\), it follows
  \begin{equation*}
      \lambda_2 \leq \frac{1}{\gamma} + \frac{\mu}{\gamma^2}L_y^2 - \frac{\mu}{1 + \frac{\mu}{\gamma}L_y^2}.
  \end{equation*}
  Finally, 
  \[
    \lambda_{\max}(A) \leq \max
    \left(\frac{1}{\gamma}, \frac{1}{\gamma} + \frac{\mu}{\gamma^2}L_y^2 - \frac{\mu}{1 + \frac{\mu}{\gamma}L_y^2}\right),
  \]
  which establishes the upper bound.
  \qed
\end{proof}



\begin{proof}[of \Cref{thm:new-L-BFGS-update}]
Consider one damped BFGS update using \(s\) and \(y\) defined in~\eqref{eq:define-y-s} and \(\hat{y}_k\) defined in~\eqref{eq:new-ytilde}, i.e, \(p = 1\),

\begin{equation*}
H_{k+1} = \hat{V}_k H_{k+1}^0 \hat{V}_k^\top  + \hat{\rho}_k s_k s_k^\top , \quad \text{where} \quad \hat{\rho}_k = 1/s_k^\top  \hat{y}_k, \quad \hat{V}_k = I-\hat{\rho}_k s_k \hat{y}_k^\top.
\end{equation*}

Let \(0 < \mu_1 := \lambda_{\min}(H_{k+1}^0) \leq \mu_2 := \lambda_{\max}(H_{k+1}^0)\).
We have
\begin{equation}
    \label{eq:eigs-update1}
    \lambda_{\min}( \mu_1 \hat{V}_k \hat{V}_k^\top  + \hat{\rho}_k s_k s_k^\top ) \leq
    \lambda_{\min}(H_{k+1}) \leq
    \lambda_{\max}(H_{k+1}) \leq \lambda_{\max}(\mu_2 \hat{V}_k \hat{V}_k^\top  +  \hat{\rho}_k s_k s_k^\top ).
\end{equation}
Let us show that we can apply \Cref{lem:eigenvalues-update} to 
\begin{equation*}
A_1 := \mu_1 \hat{V}_k \hat{V}_k^\top  + \hat{\rho}_k s_k s_k^\top  \quad \text{and} \quad A_2 := \mu_2 \hat{V}_k \hat{V}_k^\top + \hat{\rho}_k s_k s_k^\top.
\end{equation*}
From~\eqref{eq:new-sTy-bounded}, we obtain
\begin{equation*}
    s_k^\top\hat{y}_k \geq \eta \lambda_{\min}(B_{k+1}^0)\|s_k\|^2 = \frac{\eta}{\lambda_{\max}(H_{k+1}^0)}\|s_k\|^2 = \frac{\eta}{\mu_2}\|s_k\|^2. 
\end{equation*}
\Cref{asm:stochastic-g-lipschitz} yields
\begin{equation*}
    \|\hat{y}_k\| = \|\theta_k y_k + (1-\theta_k) B_{k+1}^0 s_k\|
     \leq \|y_k\|+\|B_{k+1}^0 s_k\|
     \leq (L_g + 1 / \mu_1) \|s_k\|.
\end{equation*}



Therefore, we can first apply \Cref{lem:eigenvalues-update} with \(s_k\), \(\hat{y}_k\), \(\gamma = \eta / \mu_2 > 0\), \(L_y = L_g + 1 / \mu_1 > 0\) and \(\mu = \mu_1 > 0\) for \(A_1\), and apply it again with \(\mu = \mu_2 > 0\) for \(A_2\).
Let \(L_1 := L_g + 1 / \mu_1\).
\Cref{lem:eigenvalues-update} and~\eqref{eq:eigs-update1} yield
  \begin{align*}
    \lambda_{\min}(H_{k+1}) & \geq \min \left( \frac{1}{L_1}, \frac{\mu_1} {1 + \frac{ \mu_1 \mu_2 }{\eta} L_1^2 } \right) > 0, \\
    \lambda_{\max}(H_{k+1}) &\leq \frac{\mu_2}{\eta} + \max \left( 0, \frac{ \mu_2^3 }{ \eta^2} L_1^2 - \frac{ \mu_2 }{ 1 + \frac{\mu_2^2}{\eta} L_1^2 }  \right).
  \end{align*}

Now, consider the case where \(p>1\) and let

\begin{equation*}
H_{k+1}^{(h+1)} := \hat{V}_{k-h} H_{k+1}^{(h)} \hat{V}_{k-h}^\top + \hat{\rho}_{k-h} s_{k-h} s_{k-h}^\top,  \quad 0 \leq h \leq p - 1 ,
\end{equation*}

where

\begin{equation*}
H^{(p)}_{k+1} := H_{k+1}, \quad \hat{\rho}_{k-h} = 1 / s_{k-h}^\top \hat{y}_{k-h}, \quad \hat{V}_k = I-\hat{\rho}_{k-h} s_{k-h} \hat{y}_{k-h}^\top.
\end{equation*}

Similarly to the case \(p=1\), 
we may write
\begin{alignat*}{2}
    \lambda_{\min}(H_{k+1}^{(h+1)}) &\geq \lambda_{\min}( \mu_1^{(h)} \hat{V}_{k-h} \hat{V}_{k-h}^\top + \hat{\rho}_{k-h} s_{k-h} s_{k-h}^\top), \qquad \mu_1^{(h)} & := \lambda_{\min}(H_{k+1}^{(h)}), \\
    \lambda_{\max}(H_{k+1}^{(h+1)}) &\leq \lambda_{\max}(\mu_2^{(h)} \hat{V}_{k-h} \hat{V}_{k-h}^\top +  \hat{\rho}_{k-h}s_{k-h}s_{k-h}^\top), \qquad   \mu_2^{(h)} & := \lambda_{\max}(H_{k+1}^{(h)}).
  \end{alignat*}
  
Assume by recurrence that \(0 < \mu_1^{(h)} \leq \mu_2^{(h)}\).
We show that we can apply \Cref{lem:eigenvalues-update} to 

\begin{equation*}
  A_1^{(h)} := \mu_1^{(h)} \hat{V}_{k-h} \hat{V}_{k-h}^\top + \hat{\rho}_{k-h}s_{k-h}s_{k-h}^\top \quad \text{and} \quad A_2^{(h)} := \mu_2^{(h)} \hat{V}_{k-h} \hat{V}_{k-h}^\top +  \hat{\rho}_{k-h}s_{k-h}s_{k-h}^\top.
\end{equation*}

From~\eqref{eq:new-sTy-bounded}, we have

  \begin{equation*}
    s_{k-h}^\top\hat{y}_{k-h} \geq \eta \lambda_{\min}(B_{k-h+1}^0)\|s_{k-h}\|^2 = \frac{\eta}{\lambda_{\max}(H_{k-h+1}^0)}\|s_{k-h}\|^2.
  \end{equation*}

Using \Cref{asm:stochastic-g-lipschitz},

\begin{equation*}
    \|\hat{y}_{k-h}\| = \| \theta_{k-h} y_{k-h} + (1-\theta_{k-h}) B_{k-h+1}^0 s_{k-h}\| \leq L_g \|s_{k-h}\| + \|B_{k-h+1}^0 s_{k-h}\|,
\end{equation*}

so that

\begin{equation*}
    \|\hat{y}_{k-h}\| \leq (L_g + \frac{1}{\lambda_{\min}(H_{k-h+1}^0)}) \|s_{k-h}\|.
\end{equation*}

We first apply \Cref{lem:eigenvalues-update} with \(s = s_{k-h}\), \(y = \hat{y}_{k-h}\), \(\gamma = \eta / \lambda_{\max}(H_{k-h+1}^0) > 0\), \(L_y = L_g + 1 / \lambda_{\min}(H_{k-h+1}^0) > 0\) and \(\mu = \mu_1^{(h)} > 0\) for \(A_1^{(h)}\), and apply it a second time with \(\mu = \mu_2^{(h)} > 0\) for \(A_2^{(h)}\). 
Let \(L_{k-h+1} := L_g + 1 / \lambda_{\min}(H_{k-h+1}^0)\) and \(\gamma_{k-h+1} = \eta / \lambda_{\max}(H_{k-h+1}^0)\).
Then we have

\begin{equation}
\label{eq:lambdaminpr}
\lambda_{\min}(H_{k+1}^{(h+1)}) \geq \min \left( \frac{1}{L_{k-h+1}}, \frac{ \lambda_{\min}(H_{k+1}^{(h)})} {1 + \frac{ \lambda_{\min}(H_{k+1}^{(h)})}{\gamma_{k-h+1}} L_{k-h+1}^2 } \right),
\end{equation}

and

\begin{equation}
\label{eq:lambdamaxpr}
\lambda_{\max}(H_{k+1}^{(h+1)}) \leq \frac{1}{\gamma_{k-h+1}} + \max \left( 0, \frac{ \lambda_{\max}(H_{k+1}^{(h)}) }{ \gamma_{k-h+1}^2} L_{k-h+1}^2 - \frac{ \lambda_{\max}(H_{k+1}^{(h)}) }{ 1 + \frac{\lambda_{\max}(H_{k+1}^{(h)})}{\gamma_{k-h+1}} L_{k-h+1}^2 }  \right).
\end{equation}
  
It is clear that we can obtain the lower bound on \(\lambda_{\min}(H_{k+1})\) recursively using~\eqref{eq:lambdaminpr}. Obtaining the upper bound on \(\lambda_{\max}(H_{k+1})\) using~\eqref{eq:lambdamaxpr} is trickier. However, we notice that inequality~\eqref{eq:lambdamaxpr} implies


\begin{equation}
\label{eq:lambdamaxpr2}
\lambda_{\max}(H_{k+1}^{(h+1)}) \leq \frac{1}{\gamma_{k-h+1}} + \max \left( 0, \frac{ \lambda_{\max}(H_{k+1}^{(h)}) }{ \gamma_{k-h+1}^2} L_{k-h+1}^2 - \frac{ \lambda_{\min}(H_{k+1}^{(h)}) }{ 1 + \frac{\lambda_{\max}(H_{k+1}^{(h)})}{\gamma_{k-h+1}} L_{k-h+1}^2 }  \right).
\end{equation}
This upper bound is less tight but it allows us to bound \(\lambda_{\max}(H_{k+1})\) recursively.\qed
\end{proof}

\section{Experimental details}

For second-order methods: 

\begin{itemize}
    \item We apply our Hessian norm control using the bound on the maximum and the minimum eigenvalues of \(H_{k+1}\), where the latter is equivalent to controlling \(\|B_{k+1}\|\);
    \item In the definition of \(H_{k+1}^0\) in~\eqref{eq:init_HK}, we choose \(\underline{\gamma}_{k+1}\) and \(\overline{\gamma}_{k+1}\) constant;
    \item The numerical values for all algorithms are the ones that yielded the best results among all sets of values that we experimented with. 
\end{itemize}

Numerical values:
\begin{itemize}
    \item For all algorithms: we train the network for \(20\) epochs and use a batch size of \(256\) samples;
    \item For SVRG, we choose a step size equal to \(0.001\);
    \item For both SdLBFGS-VR and \Cref{alg:VR-SdLBFGS-CHN}, the memory parameter \(p = 10\), the minimal scaling parameter \(\underline{\gamma}_{k+1} = 0.1\) for all \(k\), the constant step size \(\alpha_k = 0.1\) and \(\eta = 0.25\);
    \item For \Cref{alg:VR-SdLBFGS-CHN}, we use a maximal scaling parameter \(\overline{\gamma}_{k+1} = 10^5\) for all \(k\), a lower bound limit \(\lambda_{\min} = 10^{-5}\) and an upper bound limit \(\lambda_{\max} = 10^{5}\).
\end{itemize}


\end{document}